\documentclass[sigconf]{acmart}
\usepackage{amsmath}
\usepackage{booktabs}
\usepackage{graphicx}
\graphicspath{{figures/}{figures/b1/}}
\usepackage{subcaption}
\usepackage{multirow}
\usepackage{algorithm}
\usepackage{algpseudocode}
\usepackage{microtype}
\usepackage{array}
\usepackage{tabularx}
\usepackage{enumitem}

\usepackage[table]{xcolor}
\usepackage[dvipsnames]{xcolor}
\newcommand{\method}{EGPO}
\newcommand{\ThinkOpen}{{\textit{<think>}}}
\newcommand{\ThinkClose}{{\textit{</think>}}}

\AtBeginDocument{%
  }

\setcopyright{acmlicensed}
\copyrightyear{2018}
\acmYear{2018}
\acmDOI{XXXXXXX.XXXXXXX}
\acmConference[Conference acronym 'XX]{Make sure to enter the correct
  conference title from your rights confirmation email}{June 03--05,
  2018}{Woodstock, NY}
\acmISBN{978-1-4503-XXXX-X/2018/06}

\begin{document}

\title{Know What You Know: Metacognitive Entropy Calibration for Verifiable RL Reasoning}

\author{Qiannian Zhao}
\affiliation{%
  \institution{ }
  \country{ }}
\email{zqn200285@gmail.com}
\author{Chen Yang}
\authornote{Contributing equally with the first author.}
\affiliation{%
  \institution{ }
    \country{ }}
\email{yangchenwww@gmail.com}
\author{Jinhao Jing}
\authornotemark[1]
\affiliation{%
  \institution{ }
   \country{ }}
\email{kawakamitomie56@gmail.com}
\author{Yunke Zhang}
\affiliation{%
  \institution{ }
   \country{ }}
\email{zhangyunke1995@gmail.com}
\author{Xuhui Ren}
\authornote{Corresponding author.}
\affiliation{%
  \institution{ }
   \country{ }}
\email{phoneees@163.com}
\author{Lu Yu}
\authornotemark[2]
\affiliation{%
  \institution{ }
   \country{ }}
\email{
lu.yu@cityu.edu.hk}
\author{Shijie Zhang}
\authornotemark[2]
\affiliation{%
  \institution{ }
   \country{ }}
\email{zhang.shijie1101@gmail.com}
\author{Hongzhi Yin}
\authornotemark[2]
\affiliation{%
  \institution{ }
   \country{ }}
\email{h.yin1@uq.edu.au}

\begin{abstract}
Large reasoning models (LRMs) have emerged as a powerful paradigm for solving complex real-world tasks. In practice, these models are predominantly trained via Reinforcement Learning with Verifiable Rewards (RLVR), yet most existing outcome-only RLVR pipelines rely almost exclusively on a binary correctness signal and largely ignore the model’s intrinsic uncertainty. We term this discrepancy the uncertainty–reward mismatch, under which high- and low-uncertainty solutions are treated equivalently, preventing the policy from ``Know What You Know'' and impeding the shift from optimizing for correct answers to optimizing effective reasoning paths. This limitation is especially critical in reasoning-centric tasks such as mathematics and question answering, where performance hinges on the quality of the model’s internal reasoning process rather than mere memorization of final answers. To address this, we propose \method{}, a metacognitive entropy calibration framework that explicitly integrates intrinsic uncertainty into RLVR for enhancing LRMs. \method{} estimates per-sample uncertainty using a zero-overhead entropy proxy derived from token-level likelihoods and aligns it with extrinsic correctness through an asymmetric calibration mechanism that preserves correct reasoning while selectively regulating overconfident failures, thereby enabling stable and uncertainty-aware policy optimization. Moreover, \method{} recovers informative learning signals from otherwise degenerate group-based rollouts without modifying the verifier or reward definition. Extensive experiments across multiple benchmarks demonstrate that the proposed \method{} leads to substantial and consistent improvements in reasoning performance, establishing a principled path for advancing LRMs through metacognitive entropy calibration.
\end{abstract}

\ccsdesc[500]{Computing methodologies~Reinforcement learning}
\ccsdesc[500]{Computing methodologies~Natural language processing}
\ccsdesc[500]{Computing methodologies~Knowledge representation and reasoning}
\ccsdesc[500]{Computing methodologies~Artificial intelligence}

\keywords{Reasoning Language Models, Reinforcement Learning with Verifiable Rewards, Metacognitive entropy calibration}

\maketitle

\section{Introduction}
\label{sec:intro}

Large reasoning models (LRMs) have demonstrated remarkable capabilities in tackling complex tasks, ranging from mathematical problem solving \cite{deepseekr1, jaech2024openai, shao2024deepseekmath} to agentic decision-making \cite{wu2025agentic, chen2025learning}. Representative systems such as DeepSeek-R1 \cite{deepseekr1} and OpenAI o1 \cite{jaech2024openai} achieve remarkable performance on rigorous reasoning benchmarks, highlighting the pivotal role of explicit reasoning in modern large language models (LLMs). 
A key mechanism behind these advances is Reinforcement Learning with Verifiable Rewards (RLVR) \cite{lightman2023verify, shao2024deepseekmath, chen2026dragrpogrponeedsknow, liu2025drgrpo}, a post-training paradigm that substitutes subjective preference annotations with programmatic verifiers that return a binary reward signal based on objective correctness.

The fundamental principle of this paradigm, exemplified by Group Relative Policy Optimization (GRPO)~\cite{shao2024deepseekmath}, is to sample multiple rollouts for a single prompt and optimize the policy using group-relative advantages derived from within-group comparisons.
However, this reliance on intra-group contrast constitutes a significant bottleneck: if a sampled group consists entirely of correct or entirely incorrect responses (i.e., uniform rewards), the relative advantages vanish, resulting in ineffective zero-gradient updates. To mitigate this limitation, recent research has pursued three primary directions. The first focuses on data efficiency by prioritizing prompts that yield a mixture of correct and incorrect responses, as seen in DAPO~\cite{yu2025dapo}. The second line of work refines advantage estimation to produce more stable update signals, exemplified by Dr.GRPO~\cite{liu2025understanding}, DRA-GRPO~\cite{chen2026dragrpogrponeedsknow}, and GSPO~\cite{zhao2025gspo}. The third direction enhances exploration to prevent groups from converging to uniformly correct or incorrect states, as proposed in ERPO~\cite{liu2025erpo}.

Despite these improvements, existing outcome-only verifiers in RLVR provide a coarse, binary correctness signal while largely neglecting the LRM’s intrinsic uncertainty, namely its internal estimate of confidence during generation.
We term this discrepancy the \textbf{uncertainty–reward mismatch}, which prevents the policy from learning how uncertainty should meaningfully guide its reasoning behavior. To explicate this limitation, we categorize outcomes along the joint axes of correctness and uncertainty, yielding two broad classes of learning states. \textbf{Success states} correspond to correct solutions and comprise low-uncertainty correct, where the model is well-calibrated and truly knows the answer, and high-uncertainty correct, where the answer is correct but reached hesitantly or via fragile reasoning. \textbf{Error states} correspond to incorrect solutions and comprise low-uncertainty incorrect, akin to a hallucination that reflects a stable misconception, and high-uncertainty incorrect, which typically reflects exploratory attempts. Consequently, this mismatch exposes two fundamental deficiencies. First, it blurs the boundary between consolidating knowledge and preserving exploration. Because binary outcome rewards treat success states as equally desirable, while penalizing error states solutions in the same way, the model cannot reliably decide when to reinforce stable reasoning patterns versus when to sustain exploratory search. Second, it fails to extract meaningful learning signals from hard problems, particularly in entirely incorrect groups, where group-relative objectives collapse to zero gradients, forcing practitioners to either filter these hardest cases or inject additional supervision, thereby wasting valuable training signal.

Advancing the reasoning frontier of LLMs requires an RLVR training process that explicitly accounts for the policy’s intrinsic uncertainty, which can be quantified by the dispersion of its output distribution. EDGE-GRPO~\cite{zhang2025edgegrpo} is an early attempt in this direction that leverages Shannon entropy~\cite{shannon1948mathematical} to modulate update magnitude, but it relies solely on entropy for reweighting, such that correct and incorrect trajectories may receive comparably scaled updates. 
This makes it difficult to both consolidate correct reasoning and preserve exploration, as overly aggressive updates on failures can inadvertently suppress exploratory attempts. Moreover, injecting external supervision (e.g., ground-truth answers) to generate corrective rollouts for entirely incorrect groups departs from outcome-only RLVR and may introduce unstable learning signals when the policy is weak, such as cases where the final answer is correct but the reasoning trajectory remains flawed. Consequently, the central challenge in incorporating uncertainty into RLVR is to establish a principled mechanism that strengthens correct reasoning while still extracting meaningful learning signals from incorrect samples without suppression of exploration.

Inspired by metacognition \cite{martinez2006metacognition}, the human capacity to monitor internal uncertainty and regulate learning accordingly, we argue that uncertainty should play a central role in RLVR optimization. Intuitively, humans strongly reinforce solutions reached with high certainty, retain uncertain successes as promising directions for further refinement, and avoid overly penalizing uncertain failures that reflect exploratory reasoning. Translating this principle to RLVR, uncertainty should serve as a modulating signal for policy updates. For success states, it reflects evidential strength and determines the degree of knowledge consolidation. For error states, it acts as a damping factor that tempers updates, enabling the model to learn from failures while preserving exploratory behavior essential for discovering improved reasoning strategies.

\begin{figure*}[t]
    \centering
    \includegraphics[width=1.0\textwidth]{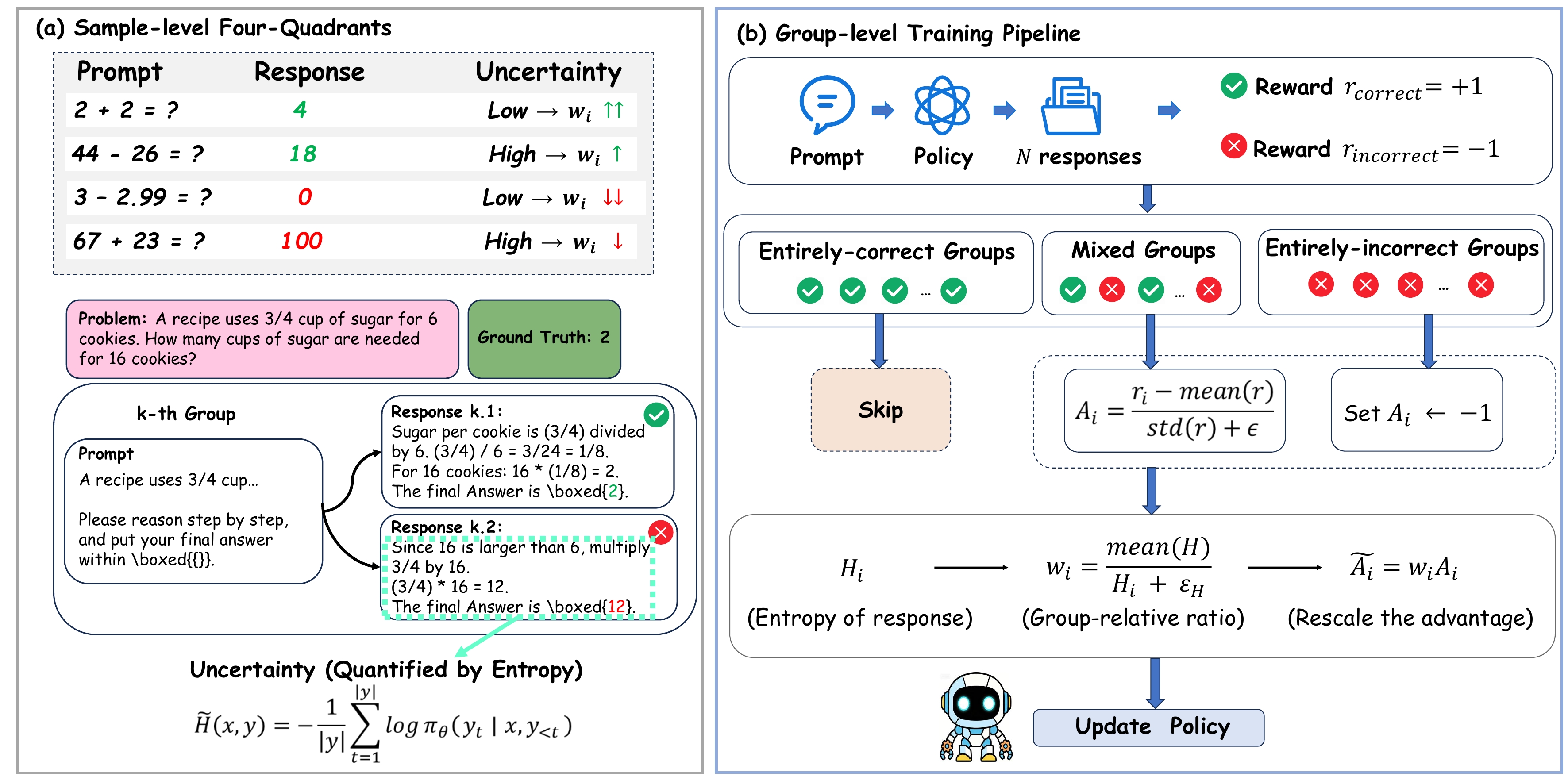}
    \caption{Overview of \method{}: sample-level metacognitive calibration (four quadrants) and group-level rollout triage (entirely-correct / mixed (contains both correct and incorrect outcomes) / entirely-incorrect).}
    \label{fig:intro_quadrants}
\end{figure*}
To this end, we propose \method{}, a lightweight \underline{\textbf{E}}ntropy-\underline{\textbf{G}}uided \underline{\textbf{P}}olicy \underline{\textbf{O}}ptimization for RLVR that injects metacognitive calibration without changing the binary outcome reward. Instead of redesigning rewards, \method{} rescales the \emph{advantage magnitude} using an intrinsic uncertainty proxy derived from the policy itself. Specifically, within each group we estimate per-sample uncertainty via sequence negative log-likelihood (NLL) \cite{shannon1948mathematical}, a cheap statistic directly tied to the policy’s token log-probabilities, and normalize it into a group-level uncertainty weight so that the overall step size remains scale-preserving while credit is redistributed within the group. We further adopt an asymmetric calibration rule: correct trajectories are never down-weighted while incorrect trajectories are never up-weighted , whereas incorrect trajectories are never up-weighted, which improves stability under sparse binary rewards. Moreover, when all rollouts in a group are incorrect and GRPO yields zero advantage, \method{} applies entropy-damped negative sample reinforcement (NSR) \cite{zhu2025nsr} to recover a non-zero learning signal while still preserving exploratory behavior. We summarize our main contributions as follows:

\begin{itemize}
    \item We identify a fundamental uncertainty–reward mismatch in outcome-only RLVR, where binary correctness signals are misaligned with the policy’s intrinsic uncertainty, leading to flawed credit assignment and wasted learning signals on hard problems. To address this, we frame the issue from a metacognitive perspective, arguing that effective RLVR should jointly consider correctness and uncertainty rather than optimizing answers alone.
    \item We propose EGPO, a lightweight verifier-agnostic training algorithm for RLVR that integrates metacognitive entropy calibration into group-based optimization. EGPO rescales advantage magnitudes using a token-level entropy proxy and applies an asymmetric calibration rule, while recovering learning signals from entirely incorrect groups via entropy-damped negative sample reinforcement, thereby providing a more principled and stable training strategy.
    \item Extensive experiments across multiple in-domain and out-of-distribution reasoning benchmarks demonstrate that EGPO consistently outperforms strong RLVR baselines, validating both its effectiveness and its robust generalization ability across models and tasks.
\end{itemize}

\section{Preliminaries}
\label{sec:method:prelim}

\subsection{Problem Setup}
\label{sec:prelim:setup}

\subsubsection{Reinforcement Learning with Verifiable Rewards (RLVR)}
\label{sec:prelim:rlvr}

RLVR studies policy optimization where reward is provided by a programmatic verifier rather than human preference~\cite{lightman2023verify,shao2024deepseekmath}.
We sample a problem instance $(x,g)\sim\mathcal{D}$, where $x$ is the prompt, $g$ is the ground truth and $\mathcal{D}$ is the training dataset.
A policy $\pi_\theta(\cdot\mid x)$ generates a response $y\sim\pi_\theta(\cdot\mid x)$, and a verifier $V$ checks correctness using only the final answer extracted from $y$.
Let $\mathrm{Ans}(y)$ denote the extracted final answer.
The outcome reward is defined as
\begin{equation}
r(y)=
\begin{cases}
+1, & \mathrm{Ans}(y)=g,\\
-1, & \text{otherwise}.
\end{cases}
\label{eq:outcome_reward}
\end{equation}
Thus, outcome-only RLVR supervises the policy only through final-answer correctness relative to $g$, and does not directly score intermediate reasoning tokens.

In verifiable math and logic tasks, we follow the standard MATH evaluation protocol~\cite{hendrycks2021math,hendrycks_math_repo} by prompting the model to reason step by step and to place the final answer within \texttt{\textbackslash boxed{}}. Accordingly, $\mathrm{Ans}(y)$ extracts the content inside the last \texttt{\textbackslash boxed{}} in $y$.

\subsubsection{Group Relative Policy Optimization (GRPO)}
\label{sec:prelim:grpo}

Group Relative Policy Optimization (GRPO)~\cite{shao2024deepseekmath} is a group-based policy optimization method widely used in outcome-only RLVR.
For a problem instance $(x,g)\sim\mathcal{D}$, we sample a group of $N$ responses $\{y_i\}_{i=1}^N$ from an old policy $\pi_{\theta_{\text{old}}}(\cdot\mid x)$ and obtain outcome rewards $\{r_i\}_{i=1}^N$ via Eq.~\eqref{eq:outcome_reward}, forming one group $\mathcal{G}(x)=\{(x,y_i,r_i)\}_{i=1}^{N}$.
GRPO constructs a group-relative advantage by normalizing outcome rewards within the group:
\begin{equation}
A_i \;=\; \frac{r_i - \mathrm{mean}(r_1,\ldots,r_N)}{\mathrm{std}(r_1,\ldots,r_N)}.
\label{eq:grpo_adv}
\end{equation}
Let $\pi_{\theta}$ denote the current policy.
GRPO updates $\pi_{\theta}$ by maximizing a clipped ratio objective:
\begin{equation}
\mathcal{L}_{\mathrm{GRPO}}(\theta)
=\mathbb{E}
\Big[
\min\big(
\rho_i(\theta)\,A_i,\;
\mathrm{clip}(\rho_i(\theta),1-\epsilon,1+\epsilon)\,A_i
\big)
\Big],
\label{eq:grpo_obj}
\end{equation}
where the expectation is over $(x,g)\sim\mathcal{D}$ and groups $\{y_i\}_{i=1}^N$ sampled from $\pi_{\theta_{\text{old}}}(\cdot\mid x)$, $\epsilon$ is the clipping coefficient, and
\begin{equation}
\rho_i(\theta) \;=\; \frac{\pi_{\theta}(y_i\mid x)}{\pi_{\theta_{\text{old}}}(y_i\mid x)}.
\label{eq:pi_ratio}
\end{equation}

\paragraph{Advantage collapse.}
With outcome rewards $r_i\in\{-1,+1\}$, a group can be \emph{mixed} ($\exists\, i,j:\ r_i\neq r_j$), \emph{entirely-correct} ($\forall i,\ r_i=+1$), or \emph{entirely-incorrect} ($\forall i,\ r_i=-1$).
In the latter two cases, the group provides no within-group preference structure: $r_i=\mathrm{mean}(r_1,\ldots,r_N)$ for all $i$. 
Then Eq.~\eqref{eq:grpo_adv} is undefined due to zero standard deviation, and in practice implementations set $A_i\equiv 0$, resulting in a vanishing update signal (advantage collapse).

\subsection{Intrinsic Uncertainty and Entropy Proxy}
\label{sec:prelim:entropy}

Intrinsic uncertainty measures the dispersion of the policy's output distribution during generation.
We quantify intrinsic uncertainty using an entropy proxy computed from token log-probabilities under the old policy.
Throughout the remainder of this paper, we use the term \emph{entropy} to refer to the NLL-based entropy proxy.

For a prompt $x$ and a response $y=(y_1,\ldots,y_T)$ with length $T$, the entropy proxy under the old policy is
\begin{equation}
\tilde{H}(x,y)
= -\frac{1}{T}\sum_{t=1}^{T}\log \pi_{\theta_{\text{old}}}\big(y_t \mid x, y_{<t}\big).
\label{eq:nll_entropy}
\end{equation}
Eq.~\eqref{eq:nll_entropy} is the per-token averaged negative log-likelihood of the sampled response under $\pi_{\theta_{\text{old}}}$: larger $\tilde{H}(x,y)$ indicates lower assigned probability and higher intrinsic uncertainty, while smaller $\tilde{H}(x,y)$ indicates higher assigned probability and lower intrinsic uncertainty.
This proxy introduces negligible overhead in RL training because the token log-probabilities $\log \pi_{\theta_{\text{old}}}(y_t\mid x,y_{<t})$ are already computed during rollout generation.

\begin{algorithm}[t]
\caption{\method{} update for one group $\mathcal{G}(x)$.}
\label{alg:egpo}
\begin{algorithmic}[1]
\Statex \textbf{Input:} group $\mathcal{G}(x)=\{(x,y_i,r_i)\}_{i=1}^{N}$ sampled by the old policy $\pi_{\theta_{\text{old}}}$; current policy $\pi_{\theta}$; clip $\epsilon$;
weight bounds $\lambda_{\min},\lambda_{\max}$; $\varepsilon_H$; \textsc{Renorm}.
\Statex \textbf{Output:} updated policy parameters $\theta$.

\State Compute entropy for each response: $\tilde{H}_i \gets \tilde{H}(x,y_i)$ by Eq.~\eqref{eq:nll_entropy}.
\State Compute group mean entropy $\bar{H}\gets \frac{1}{N}\sum_{i=1}^{N}\tilde{H}_i$.
\State Compute raw ratio weight $\hat{w}_i \gets \frac{\bar{H}}{\tilde{H}_i+\varepsilon_H}$ by Eq.~\eqref{eq:inv_ratio}.
\State Apply asymmetric clamp to obtain $w_i$ by Eq.~\eqref{eq:asymmetric_weight}.
\If{\textsc{Renorm} $=1$}
    \State Renormalize weights by Eq.~\eqref{eq:renorm}.
\EndIf

\Statex \textbf{Group-type handling:}
\If{$\forall i,\ r_i=+1$}
    \State \textbf{return} \Comment{entirely-correct: skip}
\ElsIf{$\forall i,\ r_i=-1$}
    \State Set $A_i \gets -1$ for all $i$ \Comment{entirely-incorrect: NSR}
\Else
    \State Compute GRPO advantages $A_i$ by Eq.~\eqref{eq:grpo_adv} \Comment{mixed group}
\EndIf

\State Form calibrated advantages $\tilde{A}_i \gets w_i \cdot A_i$ by Eq.~\eqref{eq:adv_egpo}.
\State Update $\theta$ by maximizing Eq.~\eqref{eq:egpo} with clip $\epsilon$.

\end{algorithmic}
\end{algorithm}

\section{Methodology}
\label{sec:method:egpo}
Building on the outcome-only RLVR setup in Sec.~\ref{sec:prelim:rlvr} and the intrinsic uncertainty measure in Sec.~\ref{sec:prelim:entropy}, we now introduce our Entropy-Guided Policy Optimization framework, \method{}.
Existing outcome-only RLVR pipelines typically rely on group-relative objectives such as GRPO to derive policy gradients.
Under binary outcome rewards, however, these objectives exhibit two structural limitations:
(i) they ignore the large variation in the policy's intrinsic uncertainty across rollouts, and
(ii) they suffer from advantage collapse on all-same groups, discarding precisely the hard prompts where the learning signal is scarce.

\method{} addresses these limitations through a \emph{metacognitive calibration} mechanism that explicitly couples outcome feedback with intrinsic uncertainty.
Rather than changing the verifier or reward definition, \method{} reshapes the \emph{group-based} update by:
(1) computing an entropy-based weight for each response and
(2) forming a calibrated advantage $\tilde{A}_i = w_i A_i$ that is both outcome-aware and uncertainty-aware.
On mixed groups, $A_i$ is the GRPO advantage in Eq.~\eqref{eq:grpo_adv} and $w_i$ is obtained via an asymmetric calibration that promotes confident correct rollouts while tempering updates on uncertain or confidently wrong ones.
On entirely-incorrect groups, \method{} adopts NSR by setting $A_i=-1$ and then applies the same calibrated weighting, recovering a non-vanishing and structured learning signal under advantage collapse.
In this way, \method{} provides a unified treatment of mixed and all-incorrect groups within a single PPO-style objective.

\subsection{Entropy-Guided Policy Optimization (EGPO)}
\label{sec:method:egpo_core}

We follow the RLVR setup in Sec.~\ref{sec:prelim:rlvr}.
For each prompt $x$, we sample a group of $N$ responses $\{y_i\}_{i=1}^{N}$ from the old policy $\pi_{\theta_{\text{old}}}(\cdot\mid x)$ and obtain outcome rewards $r_i$ via Eq.~\eqref{eq:outcome_reward}, forming $\mathcal{G}(x)=\{(x,y_i,r_i)\}_{i=1}^{N}$.
We retain the PPO-style importance ratio $\rho_i(\theta)$ and clipped objective used in GRPO (Sec.~\ref{sec:prelim:grpo}); the core innovation of EGPO lies in how the group-based advantages are constructed and calibrated by entropy within each group.

\subsubsection{Calibration mechanism: group-relative entropy weighting}
\label{sec:method:weighting}
\begin{figure}[t]
\centering
\includegraphics[width=0.92\columnwidth]{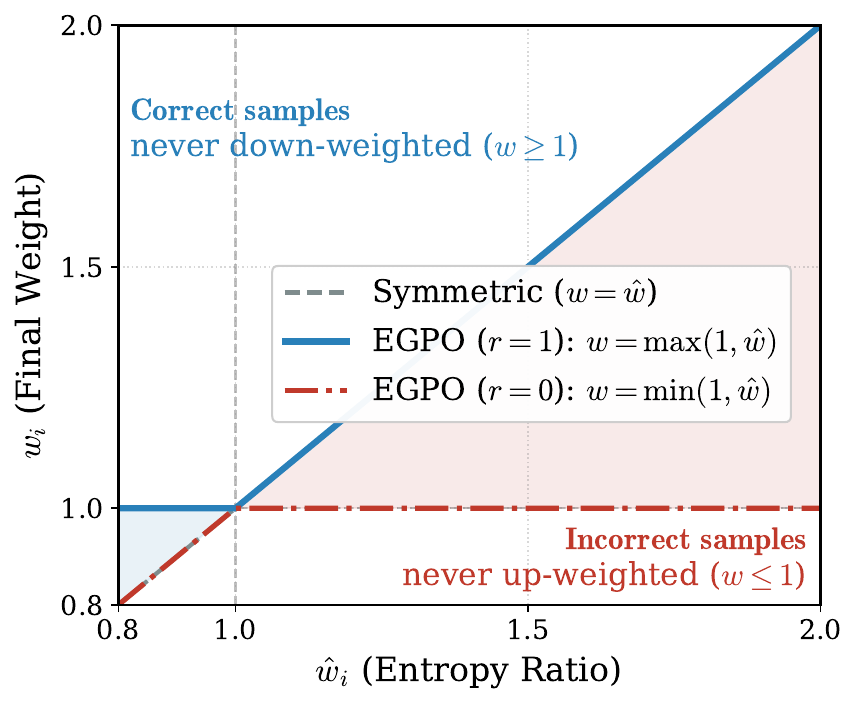}
\caption{Asymmetric calibration in \method{}: correct responses are never down-weighted and incorrect responses are never up-weighted.}
\label{fig:calib_op}
\end{figure}

We first assign an entropy value to each response using the NLL-based proxy in Eq.~\eqref{eq:nll_entropy}:
\begin{equation}
\tilde{H}_i \;\triangleq\; \tilde{H}(x,y_i).
\label{eq:egpo_entropy_i}
\end{equation}
This quantity captures how unlikely the realized trajectory $y_i$ is under $\pi_{\theta_{\text{old}}}$ and thus serves as a token-level measure of intrinsic uncertainty.

To incorporate this uncertainty in a verifier-agnostic and calibration-robust way, EGPO operates \emph{within} each group.
We compute the mean entropy $\bar{H}=\frac{1}{N}\sum_{i=1}^{N}\tilde{H}_i$ and define a group-relative weight
\begin{equation}
\hat{w}_i \;=\; \frac{\bar{H}}{\tilde{H}_i+\varepsilon_H},
\label{eq:inv_ratio}
\end{equation}
where $\varepsilon_H>0$ avoids numerical instability.
This inverse-entropy ratio has two key effects:
(i) it is scale-preserving at the group level, and
(ii) it reallocates update magnitudes among responses for the same prompt according to their \emph{relative} intrinsic uncertainty, without requiring globally calibrated entropy across prompts.

\subsubsection{Calibration mechanism: outcome-aware asymmetric calibration}
\label{sec:method:asym}

To implement the metacognitive principle in outcome-only RLVR, \method{} couples intrinsic uncertainty with verifier outcomes when scaling update.
Intuitively, high-confidence correct responses should be reinforced more aggressively, while penalties on failures should be regulated so as not to destroy latent reasoning competence or exploration.
Concretely, \method{} transforms $\hat{w}_i$ via an outcome-aware asymmetric clamp (Fig.~\ref{fig:calib_op}) with bounds $\lambda_{\min}$ and $\lambda_{\max}$:
\begin{equation}
w_i \;=\;
\begin{cases}
\max\!\Big(1.0,\,\mathrm{clip}(\hat{w}_i,\lambda_{\min},\lambda_{\max})\Big), & r_i=+1,\\
\min\!\Big(1.0,\,\mathrm{clip}(\hat{w}_i,\lambda_{\min},\lambda_{\max})\Big), & r_i=-1,
\end{cases}
\label{eq:asymmetric_weight}
\end{equation}
which enforces that correct responses are never down-weighted ($w_i\ge 1$) while incorrect responses are never up-weighted ($w_i\le 1$).
This asymmetric calibration yields a principled bias: it amplifies updates on confident correct rollouts, while ensuring that uncertainty does not translate into disproportionately large negative gradients on failures.

Optionally, \method{} applies within-group renormalization
\begin{equation}
w_i \leftarrow \frac{w_i}{\frac{1}{N}\sum_{j=1}^{N} w_j},
\label{eq:renorm}
\end{equation}
to keep the mean weight equal to $1$.
As shown in our ablation study (Sec.~\ref{sec:exp:ablation}), this renormalization can increase the performance ceiling on some benchmarks without introducing instability.

\subsubsection{Base advantage, calibrated advantage, and objective}
\label{sec:method:objective}

We now specify the base advantage $A_i$ according to the outcome pattern of $\mathcal{G}(x)$:
\begin{equation}
A_i \;=\;
\begin{cases}
0, & \textit{entirely-correct group},\\[2pt]
\frac{r_i - \mathrm{mean}(r_1,\ldots,r_N)}{\mathrm{std}(r_1,\ldots,r_N)}, & \textit{mixed group},\\[6pt]
-1, & \textit{entirely-incorrect group}.
\end{cases}
\label{eq:base_adv_cases}
\end{equation}
The mixed-group case recovers the GRPO advantage in Eq.~\eqref{eq:grpo_adv}. 
The entirely-correct case corresponds to skipping updates.
The entirely-incorrect case instantiates NSR (Sec.~\ref{sec:method:nsr}) via a constant negative advantage.

Given $A_i$ and $w_i$, EGPO forms the calibrated advantage
\begin{equation}
\tilde{A}_i \;=\; w_i \cdot A_i,
\label{eq:adv_egpo}
\end{equation}
and optimizes a weighted clipped-ratio objective:
\begin{equation}
\mathcal{L}_{\method}(\theta)=
\mathbb{E}_{(x,y_i)\sim\mathcal{G}}
\Big[
\min\big(
\rho_i(\theta)\,\tilde{A}_i,\;
\mathrm{clip}(\rho_i(\theta),1-\epsilon,1+\epsilon)\,\tilde{A}_i
\big)
\Big],
\label{eq:egpo}
\end{equation}
where $\rho_i(\theta)$ is defined in Eq.~\eqref{eq:pi_ratio} and $\epsilon$ is the PPO clipping coefficient.
Algorithm~\ref{alg:egpo} summarizes the complete EGPO update for one trajectory set.

\subsection{Negative Sample Reinforcement (NSR) for Entirely-Incorrect Groups}
\label{sec:method:nsr}

We now detail the NSR component used within EGPO for entirely-incorrect groups.
Let $\mathcal{G}^{-}(x)=\{(x,y_i,r_i)\}_{i=1}^{N}$ denote a group where $r_i=-1$ for all $i$.
Every sampled response in $\mathcal{G}^{-}(x)$ carries a negative learning signal under the outcome reward, but GRPO alone would collapse these signals to zero.

Within EGPO, we instantiate NSR by using a constant negative base advantage:
\begin{equation}
A_i \;=\; -1,\quad \forall i\in\{1,\ldots,N\}\,,
\label{eq:nsr_adv}
\end{equation}
where $-1$ serves as a normalized choice of scale and ensures a consistent negative update direction across responses.
This corresponds exactly to the \textit{entirely-incorrect} case in Eq.~\eqref{eq:base_adv_cases}.
EGPO then applies the same entropy-guided calibration as before, forming $\tilde{A}_i=w_iA_i=-w_i$ and optimizing Eq.~\eqref{eq:egpo}.

\paragraph{Effect under clipping.}
On $\mathcal{G}^{-}(x)$, the per-response term in Eq.~\eqref{eq:egpo} becomes
\begin{equation}
\begin{aligned}
\ell_i(\theta)
&= \min\Big(
    \rho_i(\theta)\,(-w_i),\;
    \mathrm{clip}(\rho_i(\theta),1-\epsilon,1+\epsilon)\,(-w_i)
  \Big) \\
&\qquad = -w_i\,\max\Big(
    \rho_i(\theta),\;
    \mathrm{clip}(\rho_i(\theta),1-\epsilon,1+\epsilon)
  \Big).
\end{aligned}
\label{eq:egpo_nsr_term}
\end{equation}
Therefore, the objective exhibits the following behavior:
(i) if $\rho_i(\theta)<1-\epsilon$, then $\ell_i(\theta)=-w_i(1-\epsilon)$ is constant in $\theta$ and contributes zero gradient;
(ii) if $1-\epsilon\le \rho_i(\theta)\le 1+\epsilon$, then $\ell_i(\theta)=-w_i\rho_i(\theta)$;
(iii) if $\rho_i(\theta)>1+\epsilon$, then $\ell_i(\theta)=-w_i\rho_i(\theta)$.
Thus, on entirely-incorrect groups with $\tilde{A}_i=-w_i<0$, clipping is active only in the low-ratio region $\rho_i(\theta)<1-\epsilon$; otherwise the objective reduces to $-w_i\rho_i(\theta)$.
Maximizing $-w_i\rho_i(\theta)$ drives $\rho_i(\theta)$ downward, which reduces the probability assigned to the sampled response under the current policy in a way that is modulated by entropy.

\paragraph{Gradient characterization.}
Consider a token step $t$ in response $y_i$.
Let $z_v$ be the logit of token $v$ and let $\pi_v$ be its softmax probability under $\pi_\theta(\cdot\mid x,y_{<t})$.
In the unclipped region, the gradient ascent direction on logits satisfies
\begin{equation}
\frac{\partial \ell_i(\theta)}{\partial z_v}
\;\propto\;
-w_i\,\rho_i(\theta)\big(\mathbf{1}[v=y_t]-\pi_v\big).
\label{eq:nsr_grad}
\end{equation}
Eq.~\eqref{eq:nsr_grad} implies that the sampled token $y_t$ is pushed down while probability mass is redistributed to other tokens at the same step, yielding a consistent negative learning signal on entirely-incorrect groups whose strength is calibrated by entropy.
The detailed derivation of Eq.~\eqref{eq:egpo_nsr_term}--Eq.~\eqref{eq:nsr_grad}, including intermediate steps, is provided in Appendix~\ref{app:nsr_derivation}.

\begin{table*}[t]
\centering
\caption{The  Pass@1 Accuracy (\%) on \textbf{1.5B} and \textbf{7B} models.
Abbreviate DeepSeek-R1-Distill-Qwen-* as DeepSeek-R1-* .}
\label{tab:main_concat_updated}
\begin{tabular*}{\textwidth}{@{\extracolsep{\fill}} l c c c c c}
\toprule
\textbf{Model} & \textbf{MATH-500} & \textbf{AIME24} & \textbf{MMLU-STEM} & \textbf{Minerva} & \textbf{GSM8K} \\
\midrule

\multicolumn{6}{l}{\textbf{1.5B Models}} \\
\midrule

Qwen2.5-Math-1.5B (Base)
  & 23.40
  & 6.67
  & 17.44
  & 19.90
  & 58.15 \\
Qwen2.5-Math-1.5B + GRPO \cite{yu2025dapo}
  & 67.20 (+43.80)
  & 13.33 (+6.66)
  & 20.84 (+3.40)
  & 32.26 (+12.36)
  & 71.65 (+13.50) \\
Qwen2.5-Math-1.5B + DAPO \cite{yu2025dapo}
  & 69.80 (+46.40)
  & 13.33 (+6.66)
  & 22.52 (+5.08)
  & 33.09 (+13.19)
  & 72.33 (+14.18) \\
Qwen2.5-Math-1.5B + EDGE-GRPO \cite{zhang2025edgegrpo}
  & 72.00 (+48.60)
  & \textbf{16.67 (+10.00)}
  & 23.78 (+6.34)
  & 33.82 (+13.92)
  & 74.68 (+16.53) \\
\rowcolor{gray!10}
Qwen2.5-Math-1.5B + EGPO (ours)
  & \textbf{74.00 (+50.60)}
  & 13.33 (+6.66)
  & \textbf{25.69 (+8.25)}
  & \textbf{37.13 (+17.23)}
  & \textbf{79.83 (+21.68)}
  \\

\addlinespace[2pt]
\cmidrule(lr){1-6}
\addlinespace[2pt]

DeepSeek-R1-1.5B (Base)
  & 81.25
  & 26.67
  & 60.26
  & 66.91
  & 78.47 \\
DeepSeek-R1-1.5B + GRPO \cite{yu2025dapo}
  & 84.45 (+3.20)
  & 26.67 (+0.00)
  & 61.84 (+1.58)
  & 66.91 (+0.00)
  & 80.20 (+1.73) \\
DeepSeek-R1-1.5B + DAPO \cite{yu2025dapo}
  & 84.45 (+3.20)
  & 23.33 (-3.34)
  & 61.53 (+1.27)
  & 67.28 (+0.37)
  & 79.08 (+0.61) \\
DeepSeek-R1-1.5B + EDGE-GRPO \cite{zhang2025edgegrpo}
  & 84.25 (+3.00)
  & 26.67 (0.00)
  & 62.10 (+1.84)
  & 67.65 (+0.74)
  & 81.27 (+2.80) \\
\rowcolor{gray!10}
DeepSeek-R1-1.5B + EGPO (ours)
  & \textbf{87.80 (+6.55)}
  & \textbf{33.33 (+6.66)}
  & \textbf{66.29 (+6.03)}
  & \textbf{70.22 (+3.31)}
  & \textbf{83.40 (+4.93)} \\

\addlinespace[4pt]
\cmidrule(lr){1-6}
\addlinespace[2pt]

\multicolumn{6}{l}{\textbf{7B Models}} \\
\midrule

Qwen2.5-Math-7B (Base)
  & 13.80
  & 3.33
  & 24.05
  & 29.41
  & 80.44 \\
Qwen2.5-Math-7B + GRPO \cite{yu2025dapo}
  & 80.80 (+67.00)
  & 26.25 (+22.92)
  & 66.60 (+42.55)
  & 41.18 (+11.77)
  & 83.40 (+2.96) \\
Qwen2.5-Math-7B + DAPO \cite{yu2025dapo}
  & 80.25 (+66.45)
  & 27.08 (+23.75)
  & 63.72 (+39.67)
  & 40.81 (+11.40)
  & 86.43 (+5.99) \\
Qwen2.5-Math-7B + EDGE-GRPO \cite{zhang2025edgegrpo}
  & 81.50 (+67.70)
  & 30.42 (+27.09)
  & 67.08 (+43.03)
  & 41.18 (+11.77)
  & 84.91 (+4.47) \\
\rowcolor{gray!10}
Qwen2.5-Math-7B + EGPO (ours)
  & \textbf{84.60 (+70.80)}
  & \textbf{33.33 (+30.00)}
  & \textbf{70.09 (+46.04)}
  & \textbf{44.49 (+15.08)}
  & \textbf{89.84 (+9.40)} \\

\addlinespace[2pt]
\cmidrule(lr){1-6}
\addlinespace[2pt]

DeepSeek-R1-7B (Base)
  & 87.48
  & 56.67
  & 85.15
  & 77.94
  & 90.06 \\
DeepSeek-R1-7B + GRPO \cite{yu2025dapo}
  & 90.80 (+3.32)
  & 54.58 (-2.09)
  & 84.99 (-0.16)
  & 78.67 (+0.73)
  & 90.60 (+0.54) \\
DeepSeek-R1-7B + DAPO \cite{yu2025dapo}
  & 91.40 (+3.92)
  & 60.00 (+3.33)
  & 84.73 (-0.42)
  & 80.15 (+2.21)
  & 90.14 (+0.08) \\
DeepSeek-R1-7B + EDGE-GRPO \cite{zhang2025edgegrpo}
  & 90.80 (+3.32)
  & 60.00 (+3.33)
  & 87.21 (+2.06)
  & 80.88 (+2.94)
  & 91.05 (+0.99) \\
\rowcolor{gray!10}
DeepSeek-R1-7B + EGPO (ours)
  & \textbf{93.40 (+5.92)}
  & \textbf{66.67 (+10.00)}
  & \textbf{91.98 (+6.83)}
  & \textbf{83.46 (+5.52)}
  & \textbf{93.18 (+3.12)} \\

\bottomrule
\end{tabular*}
\end{table*}

\section{Experiments}
\label{sec:experiments}

In this section, we perform comprehensive experiments to evaluate the effectiveness of EGPO. We explicitly address the following research questions:
\begin{itemize}
    \item \textbf{RQ1:} To what extent does \method{} improve reasoning performance and generalization over strong RL baselines?
    \item \textbf{RQ2:} How does each component of \method{} contribute to its overall effectiveness?
    \item \textbf{RQ3:} Can visualizations provide evidence of \method{}’s effectiveness?
    \item \textbf{RQ4:} Does a more fine-grained entropy guidance further improve performance?
\end{itemize}

\subsection{Experimental Setup}
\label{sec:exp_setup}

\noindent\textbf{Models and Baselines.}
We evaluate EGPO on two backbone families: Qwen2.5-Math (1.5B/7B) and DeepSeek-R1-Distill-Qwen (1.5B/7B).\footnote{Model links: \url{https://huggingface.co/Qwen/Qwen2.5-Math-1.5B}, \\
\url{https://huggingface.co/Qwen/Qwen2.5-Math-7B}, \\
\url{https://huggingface.co/deepseek-ai/DeepSeek-R1-Distill-Qwen-1.5B},\\
\url{https://huggingface.co/deepseek-ai/DeepSeek-R1-Distill-Qwen-7B}.}
We compare EGPO against GRPO~\cite{shao2024deepseekmath}, DAPO~\cite{yu2025dapo}, and EDGE-GRPO~\cite{zhang2025edgegrpo}. All methods are trained on the default binary outcome reward in Eq.~\eqref{eq:outcome_reward} for a controlled comparison.

\noindent\textbf{Training Data.}
We train on a curated subset derived from OpenR1-math-220k~\cite{openr1}. The subset is constructed to cover a representative difficulty spectrum while reducing reward noise via filtering and deduplication (full description in Appendix~\ref{app:data}). To control a small number of extreme-length outliers and keep the prompt-length distribution well-behaved, we filter samples with prompt length $>1024$, yielding the final training dataset \textsf{OpenR1-stable-10K}. We use the default binary outcome reward in Eq.~\eqref{eq:outcome_reward}.

\noindent\textbf{Training Setup.}
We conduct training using the widely used open-source RL training \emph{verl} framework \cite{sheng2024hybridflow} for all RL experiments. 
Unless stated otherwise, we sample \(N=16\) rollouts per prompt and train for 5 epochs (about 770 steps). 
The learning rate is set to \(1\times 10^{-6}\).
For our proposed EGPO method, we implement the asymmetric weight clipping with bounds \(\lambda_{\min}=0.8\) and \(\lambda_{\max}=2.0\).
All other hyperparameters, including clipping coefficient and DAPO dynamic sampling ranges, follow the default configuration in \emph{verl}.

\noindent\textbf{Benchmarks and Evaluation.}
We evaluate on MATH-500 \cite{hendrycks2021math}, AIME 2024 \cite{aime2024i,aime2024ii}, Minerva \cite{lewkowycz2022minerva}, GSM8K \cite{cobbe2021gsm8k}, and MMLU-STEM \cite{hendrycks2021mmlu} using the widely used open-source evaluation framework \emph{evalscope} \cite{evalscope} with default templates to ensure reproducibility.\footnote{Research shows that reasoning scores can be highly sensitive to prompt templates and evaluation/seeds/decoding hyperparameters \cite{liu2025understanding}. We therefore fix the evaluation harness and use the default templates throughout for fair comparison.}
To maximize comparability and robustness, we adopt a unified decoding configuration: we set temperature $T=0.6$, top-$p=0.95$, and top-$k=20$ with random sampling enabled (do\_sample=True).
For each benchmark, we conduct 8 independent trials and report the mean  pass@1 accuracy.
The maximum generation length is set to 3,072 tokens for Qwen2.5-Math series (fitting their context constraints) and 16,384 tokens for DeepSeek-R1-Distill-Qwen to prevent truncation.

\begin{table*}[t]
\centering
\caption{Ablations on Qwen2.5-Math-1.5B under the default binary outcome reward and $w_i$ denotes EGPO's group-relative weight.}
\begin{tabularx}{1 \textwidth}{l l l l p{1.45cm}<{\centering} p{0.9cm}<{\centering} p{0.9cm}<{\centering} c}
\toprule
Variant &
Constraint on $w_i$ &
Entirely-incorrect groups &
Renorm &
MATH-500 & Minerva & GSM8K & Avg. \\
\midrule
Base &
-- &
-- &
-- &
23.4 & 19.90 & 58.15 & 33.82 \\

C1 (symmetric) &
$\lambda_{\min}\le w_i\le \lambda_{\max}$ &
NSR + $w_i$ &
none &
66.0 & 33.09 & 80.97 & 60.02 \\

C2 (neg $\le 1$ only) &
$w_i\le 1$ when $r_i=-1$ only &
NSR + $w_i$ &
none &
70.6 & 30.51 & 78.01 & 59.71 \\

C3 (pos $\ge 1$ only) &
$w_i\ge 1$ when $r_i=+1$ only &
NSR + $w_i$ &
none &
72.6 & 33.04 & 80.59 & 62.08 \\

C4 (neg locked to 1) &
asymmetric clamp &
NSR with $w_i\equiv 1$ &
none &
72.0 & 34.61 & 79.68 & 62.10 \\

C5 (clamp $\rightarrow$ renorm) &
asymmetric clamp &
NSR + $w_i$ &
after clamp  &
81.05 & 35.29 & 80.29 & 65.54 \\

C6 (renorm $\rightarrow$ clamp) &
asymmetric clamp &
NSR + $w_i$ &
before clamp &
72.2 & 34.19 & 80.06 & 62.15 \\

EGPO &
asymmetric clamp &
NSR + $w_i$ &
none &
76.40 & 37.13 & 81.43 & 65.97 \\
\bottomrule
\end{tabularx}
\label{tab:ablation}
\end{table*}
\begin{figure*}[t]
    \centering
    \includegraphics[width=1.0\textwidth]{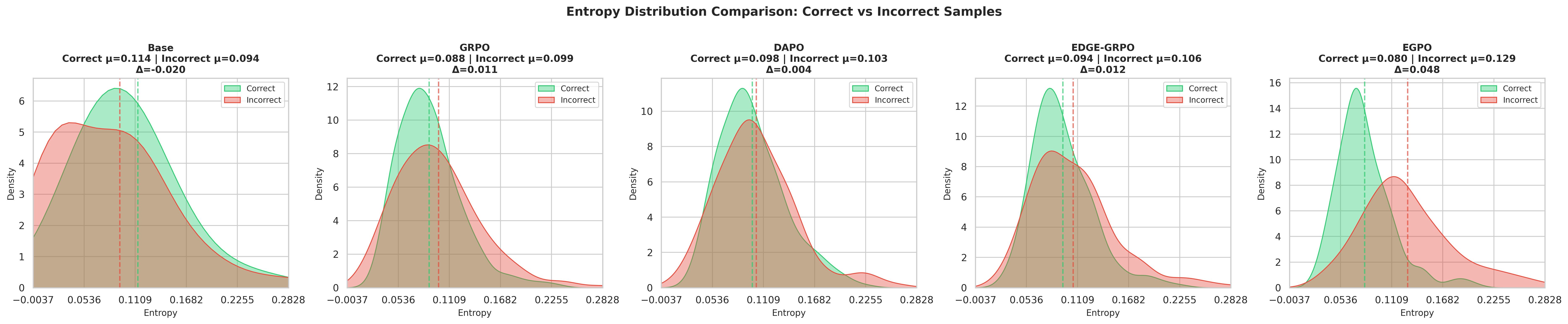}
    \caption{Density comparison of the entropy  on the response for Qwen2.5-Math-1.5B, comparing Base/GRPO/DAPO/EDGE-GRPO/\method{} on correct vs.\ incorrect rollouts.}
    \label{fig:rq1_entropy_density}
\end{figure*}

\subsection{RQ1: Comparison with Baselines}
\label{sec:exp:main}

Table~\ref{tab:main_concat_updated} reports the main results on four mathematics benchmarks and MMLU-STEM to evaluate both effectiveness and generalization. Across both 1.5B and 7B model scales, \method{} consistently outperforms strong RL baselines on nearly all benchmarks and achieves the best overall performance. Notably, EDGE-GRPO, as an early uncertainty-aware RLVR baseline, already surpasses most other RLVR methods in many cases, validating the importance of incorporating uncertainty into RLVR, yet \method{} still delivers consistent and clear gains over EDGE-GRPO, indicating that our metacognitive uncertainty calibration leverages uncertainty more effectively. Compared with base models such as Qwen2.5-Math and DeepSeek-R1, \method{} yields substantial improvements, particularly on Qwen2.5-Math-7B, where it boosts MATH-500 by +70.80\% over Base and further exceeds the strongest RL baseline by +3.10\%, while also achieving state-of-the-art results on Minerva and GSM8K. In addition, \method{} exhibits strong generalization on MMLU-STEM (e.g., +46.04\% over Base), suggesting that integrating intrinsic uncertainty into RLVR not only enhances in-domain reasoning performance but also strengthens generalizable reasoning behaviors beyond training domains. Moreover, performance margins tend to be larger at 7B, implying that \method{} more effectively raises the performance ceiling when the model capacity is sufficient.

AIME24 consists of high-difficulty competition problems, so all models exhibit relatively low absolute performance, largely due to the 4096 context length constraint that can truncate long derivations. With larger base model capacity, this limitation is partially alleviated and \method{} demonstrates clear advantages on hard problems. On Qwen2.5-Math-7B, \method{} improves AIME24 by +30.00\% over Base and further exceeds the strongest RL baseline EDGE-GRPO by +2.91\%. On DeepSeek-R1-7B, \method{} reaches 66.67\% and surpasses the strongest RL baselines by +6.67\%. These results indicate that \method{} effectively extracts informative learning signals even on the most challenging instances under outcome-only RLVR. A detailed per-backbone and per-scale analysis is provided in Appendix~\ref{app:rq1_details}.

\subsection{RQ2: Ablation Analysis}
\label{sec:exp:ablation}

To evaluate the contribution of each module, we conduct ablation study on Qwen2.5-Math-1.5B under the default binary outcome reward. All variants follow the same training setup and only differ in how constraints are applied to $w_i$ and how entirely-incorrect groups are handled (Table~\ref{tab:ablation}).

\subsubsection{Asymmetric constraints as a mechanism for balancing exploitation and exploration.}
EGPO enforces an outcome-aware asymmetric clamp on $w_i$ (never down-weight correct rollouts; never up-weight incorrect rollouts), which strengthens exploitation while avoiding overly aggressive negative updates that can reduce exploration. C1 instead uses a symmetric clamp $\lambda_{\min}\le w_i\le \lambda_{\max}$, allowing both down-weighting correct rollouts and up-weighting incorrect rollouts, and is consistently worse than EGPO (e.g., 66.0 vs.\ 76.40 on MATH-500; 33.09 vs.\ 37.13 on Minerva). The one-sided constraints (C2/C3) remain inferior, indicating both sides of the asymmetric clamp are needed for a better exploration--exploitation balance.

\subsubsection{Entirely-incorrect groups: extracting learning signal while preserving exploration.}
Entirely-incorrect groups require NSR to obtain learning signals. Fixing $w_i\equiv 1$ on such groups (C4) removes entropy-conditioned redistribution and applies uniform negative pressure regardless of entropy. EGPO applies NSR together with $w_i$, penalizing high-entropy failures more conservatively, which better preserves exploration and improves performance on harder prompts (e.g., Minerva: 37.13 vs.\ 34.61).

\subsubsection{Renorm: where it is applied matters.}
Applying renorm \textit{after clamp} (C5) can increase the performance ceiling on some benchmarks without introducing instability (e.g., GSM8K). In contrast, applying renorm \textit{before clamp} (C6) is consistently less effective, suggesting that re-centering weights prior to enforcing asymmetry weakens the intended redistribution of update magnitude. However, the gains from C5 are not consistent across benchmarks and do not improve the overall average compared with EGPO. Therefore, we use the asymmetric clamp without renorm as the default setting.

\subsection{RQ 3: Visualization}
Beyond pass@1, we analyze the response entropy distribution as defined in Eq.~\eqref{eq:nll_entropy} and conduct a qualitative case study to further demonstrate the advantages of \method{} over the base model. As shown in Fig.~\ref{fig:rq1_entropy_density}, \method{} achieves the most pronounced separation between correct and incorrect rollouts in terms of response entropy. In particular,\method{} produces a much lower mean entropy for correct samples ($\mu=0.080$) and a substantially higher mean entropy for incorrect samples ($\mu=0.129$), resulting in the largest mean gap ($\Delta=0.048$). In contrast, GRPO and EDGE-GRPO exhibit only modest gaps ($\Delta=0.011$ and $\Delta=0.012$), DAPO shows a marginal gap ($\Delta=0.004$), and Base even displays a reversed ordering with $\Delta=-0.020$. These results indicate that \method{} effectively aligns intrinsic uncertainty with verifier outcomes at the final answer level, making the update signal increasingly targeted under binary outcome rewards and enabling the model to realize the principle of ``Know What You Know”.

\begin{figure}[t]
    \centering
    \includegraphics[width=0.95\columnwidth]{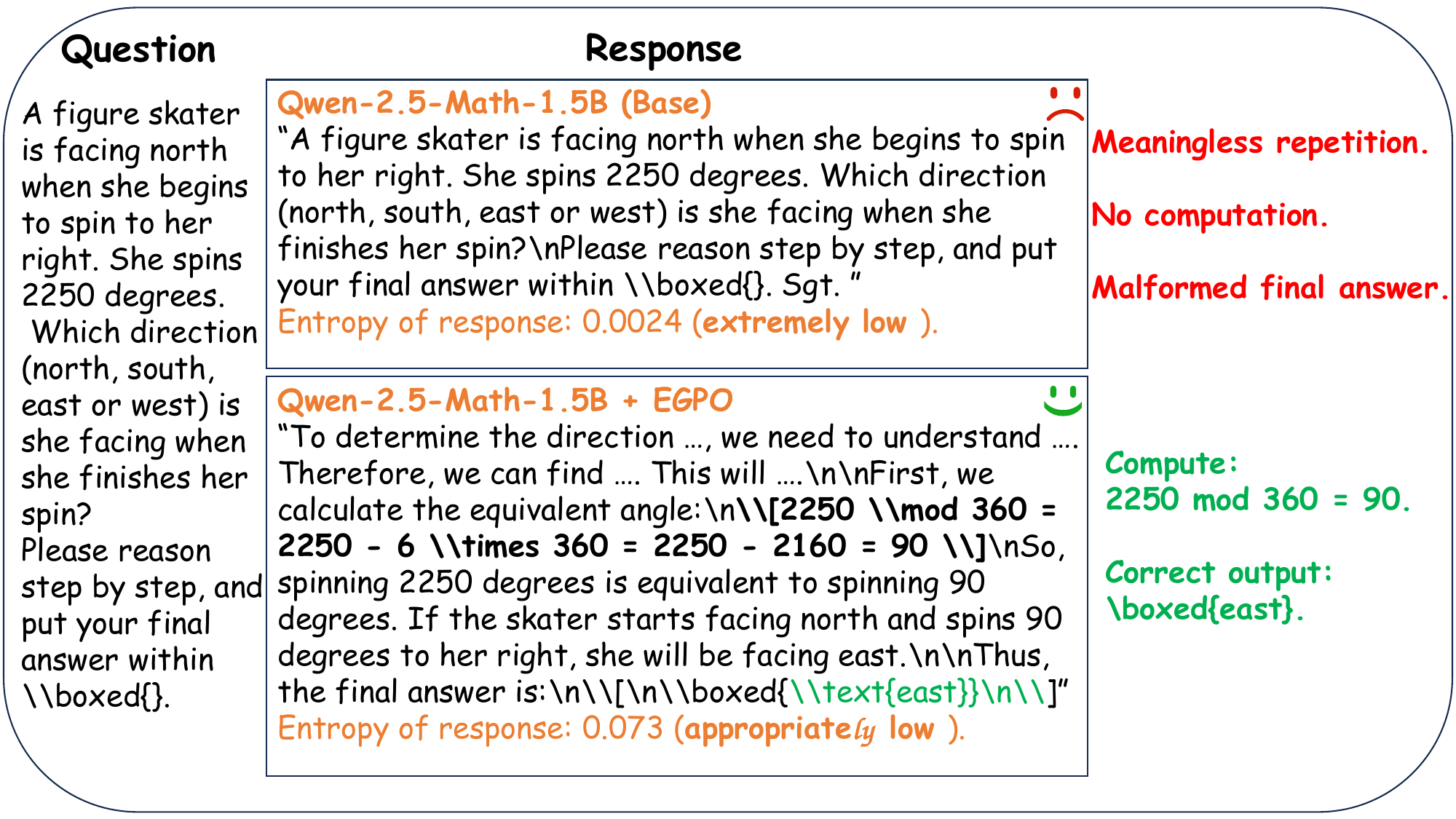}
    \caption{A qualitative case on Qwen2.5-Math-1.5B comparing Base vs.\ \method{}, including  the entropy on the response.}
    \label{fig:rq1_case_examples}
\end{figure}

To further substantiate this finding, Fig.~\ref{fig:rq1_case_examples} presents a representative case comparison between Qwen-2.5-Math-1.5B and \method{} under the same setting. In this example, \method{} exhibits higher entropy yet ultimately arrives at a correct and well-supported solution, whereas the base model fails to reach the correct answer. This qualitative evidence complements our quantitative analysis and provides additional validation of the effectiveness of \method{} in guiding more reliable and effective reasoning.
\subsection{RQ4: Extended Evaluations}
\label{sec:exp:diagnostic}

Some representative reasoning models (e.g., DeepSeek-R1) explicitly separate internal reasoning from final answers using dedicated <think>...</think> tokens. The content within <think> captures the model’s intermediate reasoning process, including tentative steps, backtracking, and self-correction, whereas the text outside <think> corresponds to the finalized answer presented to the user. Accordingly, we distinguish two types of uncertainty: \textbf{Thinking Entropy (TE)}, computed over the reasoning trace inside <think>, and \textbf{Answer Entropy (AE)}, computed over the distribution of the final answer. To provide a more comprehensive analysis of our method, we systematically evaluate both uncertainty signals. Specifically, we sample 1,000 problems and generate 8 rollouts per problem using DeepSeek-R1-Distill-Qwen backbones. As shown in Table~\ref{tab:b1_summary} , AE is consistently more predictive of incorrectness than Thinking-side entropy, with AUC improving from $0.745$ to $0.778$ on 1.5B and from $0.700$ to $0.738$ on 7B, which suggests that fine-grained entropy guidance should focus on AE to better align the entropy signal with verifier outcomes. The corresponding ROC curves are provided in Appendix~\ref{app:diagnostic} (Fig.~\ref{fig:roc_curve}), and we further confirm that this result is not driven by response length in Appendix~\ref{app:diagnostic}.

\section{Related Work}
\label{sec:related}

\textbf{Reasoning and verifiable training.}
Reasoning performance has improved via chain-of-thought (CoT) prompting, self-consistency, and search-based prompting \cite{wei2022cot,kojima2022zeroshotcot,wang2022selfconsistency,yao2023tot,yao2022react}.
RLVR leverages verifiable signals (e.g., exact answers, symbolic checking, or program execution) to scale post-training beyond human preference labels \cite{lightman2023verify,shao2024deepseekmath}.
Recent work explores richer rubric-based or multi-signal verification \cite{liu2025openrubrics,he2025beyondcorrectness,tan2025aurora} and analyzes R1-Zero style training dynamics \cite{liu2025drgrpo,chen2026dragrpogrponeedsknow}.
In contrast, \method{} keeps the verifier unchanged and focuses on calibrating the policy update magnitude.

\textbf{RLHF, PPO-style alignment, and group-based training.}
RLHF and preference learning optimize policies from comparisons, typically optimized with policy-gradient and trust-region updates \cite{williams1992reinforce,konda2000actorcritic,kakade2002natural,schulman2015trpo,schulman2016gae} \cite{christiano2017preferences,stiennon2020summarize,ziegler2019rlhf}, while DPO-style objectives avoid explicit reward modeling \cite{rafailov2023dpo}.
For verifiable tasks, group-based objectives such as GRPO reduce variance via within-group normalization and have become a common RLVR primitive \cite{shao2024deepseekmath,yu2025dapo}.
Recent variants study stability and bias (e.g., DAPO/Dr.GRPO/DRA-GRPO/GSPO) \cite{yu2025dapo,liu2025drgrpo,chen2026dragrpogrponeedsknow,zhao2025gspo} and advantage estimators such as key-token credit assignment \cite{ktae2025}.
\method{} is complementary: it modifies neither the reward nor the verifier, but rescales update magnitudes using an intrinsic uncertainty proxy and additionally recovers learning signal from entirely-incorrect groups.

\textbf{Uncertainty, calibration, and metacognition in LLMs.}
Classic calibration work studies the relationship between probabilistic uncertainty and accuracy \cite{guo2017calibration,niculescu2005predicting,ovadia2019can}.
Recent works use uncertainty to guide or stabilize training \cite{du2025confidenceasreward,yoon2025pacr,tang2025trialerror}.
AURORA aligns reward optimization with uncertainty-aware signals \cite{tan2025aurora}.
\method{} differs in mechanism: it does not add uncertainty to the reward; instead, it treats uncertainty as a metacognitive \emph{scaling factor} on policy gradients, aligning intrinsic uncertainty with extrinsic correctness.

\textbf{Entropy in RL and in LLM post-training.}
Entropy regularization is central to maximum-entropy RL \cite{ziebart2008maxentirl,haarnoja2018sac} and appears in LLM post-training as a stabilizer \cite{yu2025dapo,agarwal2025entropymin}.
However, entropy minimization alone can encourage overconfident collapse or shortcut behavior \cite{agarwal2025entropymin,cui2025entropymechanism}.
EDGE-GRPO \cite{zhang2025edgegrpo} introduced using a group-relative entropy ratio as a practical weighting construction in GRPO-style RLVR.
In this work, we treat such entropy-ratio weighting as one intrinsic metacognitive signal for uncertainty, and we directly compare against EDGE-GRPO in Sec.~\ref{sec:exp:main}.
Our focus is not to change the verifier or reward, but to make the use of uncertainty signals \emph{outcome-aware} and robust under binary RLVR, via asymmetric constraints and explicit learning from entirely-incorrect groups.

\section{Conclusion}
\label{sec:conclusion}

\begin{table}[t]
\centering
\caption{ROC-AUC for predicting incorrectness using entropy (higher is better). }
\begin{tabular}{lp{1.6cm}<{\centering}p{1.6cm}<{\centering}}
\toprule
Backbone & TE & AE\\
\midrule
DeepSeek-R1-Distill-Qwen-1.5B & 0.745 & 0.778 \\
DeepSeek-R1-Distill-Qwen-7B & 0.700 & 0.738 \\
\bottomrule
\end{tabular}

\label{tab:b1_summary}
\end{table}

In this work, we introduced \method{}, a metacognitive entropy calibration framework for outcome-only RLVR that bridges the gap between intrinsic uncertainty and extrinsic correctness. By rescaling update magnitudes within group-based policy optimization, \method{} systematically aligns entropy-based uncertainty with the default binary verifier without modifying the reward definition. Specifically, we enforce an outcome-aware asymmetric calibration rule that guarantees correct rollouts are never down-weighted while preventing incorrect rollouts from being up-weighted, thereby improving the exploration–exploitation balance under sparse verifiable rewards. Furthermore, \method{} integrates entropy-damped NSR with group-relative weighting to recover informative learning signals from entirely incorrect groups under advantage collapse, while avoiding overly aggressive suppression of uncertain failures. Empirically, across multiple benchmarks, \method{} delivers consistent and substantial gains on both Qwen2.5-Math and DeepSeek-R1-Distill-Qwen backbones, validating the effectiveness of metacognitive entropy calibration for advancing reasoning language models.

\bibliographystyle{ACM-Reference-Format}
\bibliography{sample-base}

\clearpage
\appendix

\section{Supplementary Material}
\label{app:supp}

\subsection{Derivation for NSR updates on entirely-incorrect groups}
\label{app:nsr_derivation}

We derive Eq.~\eqref{eq:nsr_grad} from the \method{} objective (Eq.~\eqref{eq:egpo}) on an entirely-incorrect group.
We consider one group $\mathcal{G}^{-}(x)=\{(x,y_i,r_i)\}_{i=1}^{N}$ with $r_i=-1$ for all $i$.
Under NSR, the base advantage is $A_i=-1$ (Eq.~\eqref{eq:nsr_adv}), and the calibrated advantage becomes
\begin{equation}
\tilde{A}_i \;=\; w_iA_i \;=\; -w_i.
\label{eq:app_nsr_calib_adv}
\end{equation}

\paragraph{Clipped objective term.}
For one sampled response $y_i$, the per-response term in Eq.~\eqref{eq:egpo} is
\begin{equation}
\ell_i(\theta)
=
\min\Big(
\rho_i(\theta)\,\tilde{A}_i,\;
\mathrm{clip}(\rho_i(\theta),1-\epsilon,1+\epsilon)\,\tilde{A}_i
\Big).
\label{eq:app_nsr_li_def}
\end{equation}

Substituting $\tilde{A}_i=-w_i$ yields
\begin{equation}
\begin{aligned}
\ell_i(\theta)
&= \min\Big(
    \rho_i(\theta)\,(-w_i),\;
    \mathrm{clip}(\rho_i(\theta),1-\epsilon,1+\epsilon)\,(-w_i)
  \Big) \\
&\qquad = -w_i\,\max\Big(
    \rho_i(\theta),\;
    \mathrm{clip}(\rho_i(\theta),1-\epsilon,1+\epsilon)
  \Big).
\end{aligned}
\label{eq:app_nsr_li_max}
\end{equation}
where the equality follows because multiplying by a negative scalar reverses the order.

\paragraph{Effect of clipping.}
Let $\rho=\rho_i(\theta)$.
By the definition of $\mathrm{clip}(\cdot)$,
\begin{equation}
\mathrm{clip}(\rho,1-\epsilon,1+\epsilon)=
\begin{cases}
1-\epsilon, & \rho<1-\epsilon,\\
\rho, & 1-\epsilon\le\rho\le 1+\epsilon,\\
1+\epsilon, & \rho>1+\epsilon.
\end{cases}
\label{eq:app_clip_cases}
\end{equation}
Plugging Eq.~\eqref{eq:app_clip_cases} into Eq.~\eqref{eq:app_nsr_li_max} gives three cases:
(i) if $\rho<1-\epsilon$, then $\max(\rho,\mathrm{clip}(\rho))=1-\epsilon$ and $\ell_i(\theta)=-w_i(1-\epsilon)$ is constant in $\rho$, thus contributing zero gradient;
(ii) if $1-\epsilon\le\rho\le 1+\epsilon$, then $\mathrm{clip}(\rho)=\rho$ and $\ell_i(\theta)=-w_i\rho$;
(iii) if $\rho>1+\epsilon$, then $\mathrm{clip}(\rho)=1+\epsilon<\rho$ and $\ell_i(\theta)=-w_i\rho$.
Therefore, for negative advantages on entirely-incorrect groups, clipping is active only when $\rho<1-\epsilon$; otherwise the objective reduces to
\begin{equation}
\ell_i(\theta) \;=\; -w_i\,\rho_i(\theta).
\label{eq:app_nsr_li_unclipped}
\end{equation}

\paragraph{Gradient direction in the unclipped region.}
We derive the gradient of Eq.~\eqref{eq:app_nsr_li_unclipped}.
Since $\rho_i(\theta)=\pi_\theta(y_i\mid x)/\pi_{\theta_{\text{old}}}(y_i\mid x)$ and $\pi_{\theta_{\text{old}}}(y_i\mid x)$ is constant w.r.t.\ $\theta$, we have
\begin{equation}
\frac{\partial \rho_i(\theta)}{\partial z_v}
=
\rho_i(\theta)\,
\frac{\partial \log \pi_\theta(y_i\mid x)}{\partial z_v}.
\label{eq:app_drho}
\end{equation}
Moreover,
\begin{equation}
\log \pi_\theta(y_i\mid x)
=
\sum_{t=1}^{T_i} \log \pi_\theta(y_t \mid x, y_{<t}).
\label{eq:app_logpi_sum}
\end{equation}
At a fixed step $t$, let $\pi_v$ be the softmax probability of token $v$ under $\pi_\theta(\cdot\mid x,y_{<t})$ and $z_v$ the corresponding logit.
The standard softmax derivative gives
\begin{equation}
\frac{\partial \log \pi_\theta(y_t \mid x, y_{<t})}{\partial z_v}
=
\mathbf{1}[v=y_t]-\pi_v.
\label{eq:app_dlogpi}
\end{equation}
Combining Eq.~\eqref{eq:app_nsr_li_unclipped}--Eq.~\eqref{eq:app_dlogpi},
\begin{equation}
\frac{\partial \ell_i(\theta)}{\partial z_v}
=
-w_i\,
\frac{\partial \rho_i(\theta)}{\partial z_v}
=
-w_i\,\rho_i(\theta)\big(\mathbf{1}[v=y_t]-\pi_v\big).
\label{eq:app_dli}
\end{equation}
Since \method{} maximizes Eq.~\eqref{eq:egpo}, the gradient ascent direction on logits is
\begin{equation}
\frac{\partial \mathcal{L}_{\method}}{\partial z_v}
\propto
w_i\,\rho_i(\theta)\big(\mathbf{1}[v=y_t]-\pi_v\big),
\label{eq:app_nsr_grad_match}
\end{equation}
which matches Eq.~\eqref{eq:nsr_grad}.
In particular, the sampled token $y_t$ is pushed down while probability mass is redistributed to other tokens, yielding a consistent negative update direction on entirely-incorrect groups.

\subsection{Dataset Construction: OpenR1-Math Stable-10k}
\label{app:data}

We load OpenR1-Math shards \cite{openr1} consisting of default and extended subsets, construct candidates with heavy filtering, apply (optional) test-set decontamination and internal deduplication via MinHash-LSH, then sample a 10k pool with subset/bucket/source constraints, and finally split out a small validation set. The goal is to produce an RL compatible pool with reduced reward noise and controlled coverage. This process yields the training dataset OpenR1-Stable-10k.

We observed that for Qwen2.5-Math (4096 context length), over 99\% of generated responses stay below 2500 tokens. Therefore, even with the longest prompts, the total context length remains well, and no special length adjustment beyond the standard prompt filtering is required.

\begin{table}[H] 
\centering
\footnotesize 
\caption{Construction recipe for the \textbf{OpenR1-Stable-10k} training dataset.}
\label{tab:dataset_recipe}
\begin{tabularx}{\columnwidth}{@{} l >{\raggedright\arraybackslash}X @{}}
\toprule
\textbf{Item} & \textbf{Setting} \\
\midrule
Pool size / Val size & 10,000 / 128 \\
Subset ratio & default 0.85, extended 0.15 \\
Bucket ratio & mixed 0.60, entirely\_correct 0.40 \newline (entirely\_incorrect retained for robustness) \\
Min generations & $\ge 2$ \\
Length filters & prompt $\le 4096$; response $\le 4096$ \newline (tokenizer-based) \\
Answer constraints & single-value GT enforced; \newline empty GT dropped \\
Boxed-match filter & enabled in Stable mode (solution or any generation last \textbackslash boxed\{\ \} matches answer) \\
Decontam / dedup & enabled (MinHash-LSH, num\_perm=128, threshold=0.8) \\
Prompt Filter ($\le1024$) & Applied: drop prompts longer than 1024 tokens (8 samples removed). \\
\bottomrule
\end{tabularx}
\end{table}

\subsection{RQ1 Detailed Analysis by Backbone and Scale}
\label{app:rq1_details}

Here provides a detailed per-backbone and per-scale comparison for RQ1.

On Qwen2.5-Math-1.5B, \method{} achieves the best results on four out of five benchmarks.
It improves MATH-500 by +50.60\% over Base and further exceeds the strongest RL baseline EDGE-GRPO by +2.00\%.
On MMLU-STEM/Minerva/GSM8K, \method{} also ranks first, surpassing EDGE-GRPO by +1.91\%, +3.31\%, and +5.15\%, respectively.
AIME24 is an exception: under the 4096 context length constraint and the 30-problem discreteness (3.33\% per solved instance), \method{} is comparable to GRPO/DAPO (13.33\%) while EDGE-GRPO solves one additional problem (16.67\%).

On Qwen2.5-Math-7B, \method{} achieves the best results on all benchmarks.
Compared to Base, it delivers large gains, including +70.80\% on MATH-500 and +46.04\% on MMLU-STEM.
It also consistently surpasses the strongest RL baselines: +3.10\% on MATH-500 (vs.\ EDGE-GRPO), +2.91\% on AIME24 (vs.\ EDGE-GRPO), +3.01\% on MMLU-STEM (vs.\ EDGE-GRPO), +3.31\% on Minerva (vs.\ GRPO/EDGE-GRPO), and +3.41\% on GSM8K (vs.\ DAPO).

On DeepSeek-R1-1.5B, \method{} achieves the best results across all benchmarks and improves over both Base and RL baselines.
It exceeds the strongest RL baselines by +3.35\% on MATH-500 (vs.\ GRPO/DAPO), +6.66\% on AIME24 (vs.\ GRPO/EDGE-GRPO), +4.19\% on MMLU-STEM (vs.\ EDGE-GRPO), +2.57\% on Minerva (vs.\ EDGE-GRPO), and +2.13\% on GSM8K (vs.\ EDGE-GRPO).

On DeepSeek-R1-7B, \method{} again achieves the best results across all benchmarks.
Notably, some RL baselines provide limited or even negative gains over Base on AIME24 and MMLU-STEM (e.g., GRPO yields -2.09\% on AIME24 and -0.16\% on MMLU-STEM), whereas \method{} improves them substantially (+10.00\% on AIME24 and +6.83\% on MMLU-STEM over Base).
Against the strongest RL baselines, \method{} further gains +2.00\% on MATH-500 (vs.\ DAPO), +6.67\% on AIME24 (vs.\ DAPO/EDGE-GRPO), +4.77\% on MMLU-STEM (vs.\ EDGE-GRPO), +2.58\% on Minerva (vs.\ EDGE-GRPO), and +2.13\% on GSM8K (vs.\ EDGE-GRPO).

\subsection{Additional Analysis: Uncertainty Diagnostics}
\label{app:diagnostic}

This subsection provides additional uncertainty diagnostics for long-trace reasoning models that expose explicit \ThinkOpen\!\dots\!\ThinkClose\ markers, enabling a clean separation between the \emph{thinking} segment and the final \emph{answer} segment.
Our analysis is conducted on the DeepSeek-R1-Distill-Qwen series.
We focus on the NLL-based entropy proxy and evaluate (i) its ability to discriminate correct vs.\ incorrect rollouts, and (ii) its robustness to response-length variations.

\begin{figure*}[!htbp]
\centering
\includegraphics[width=1\textwidth]{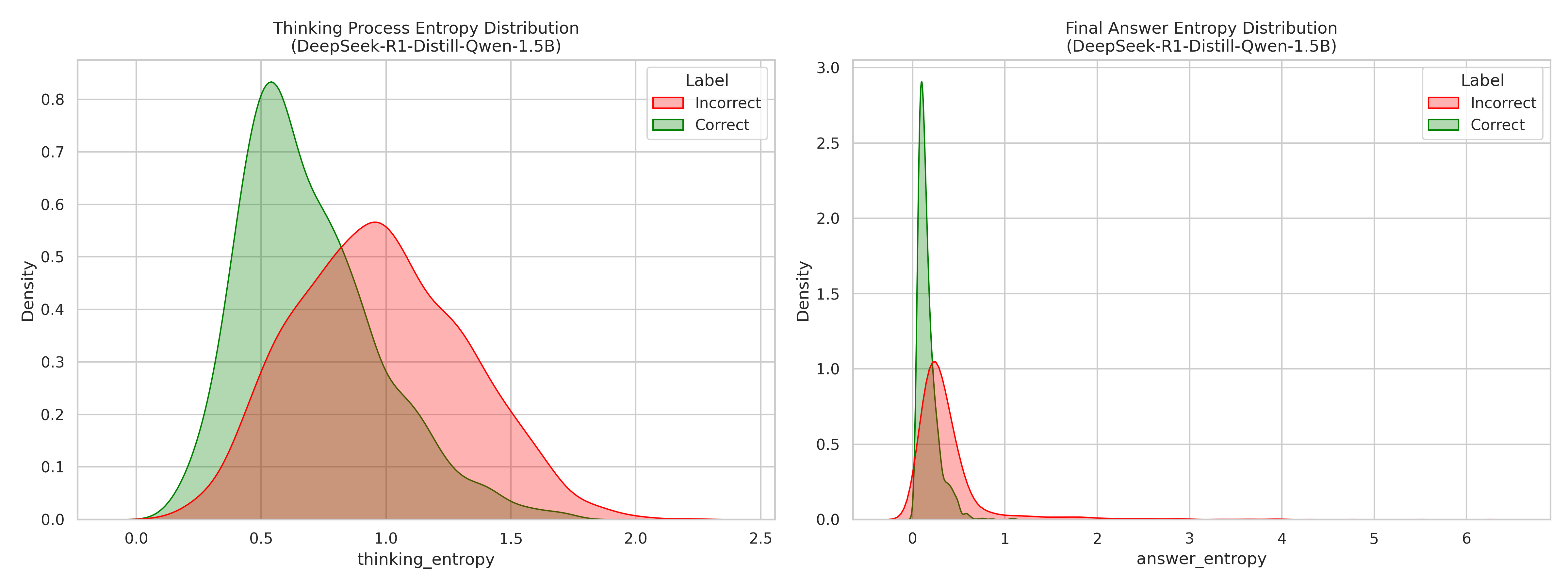}
\vspace{-1mm}
\caption{Entropy distributions for DeepSeek-R1-Distill-Qwen-1.5B (thinking-side vs.\ answer-side; correct vs.\ incorrect).}
\label{fig:app_ds_1p5b_dist}
\vspace{-2mm}
\end{figure*}

\begin{figure*}[!htbp]
\centering
\includegraphics[width=1\textwidth]{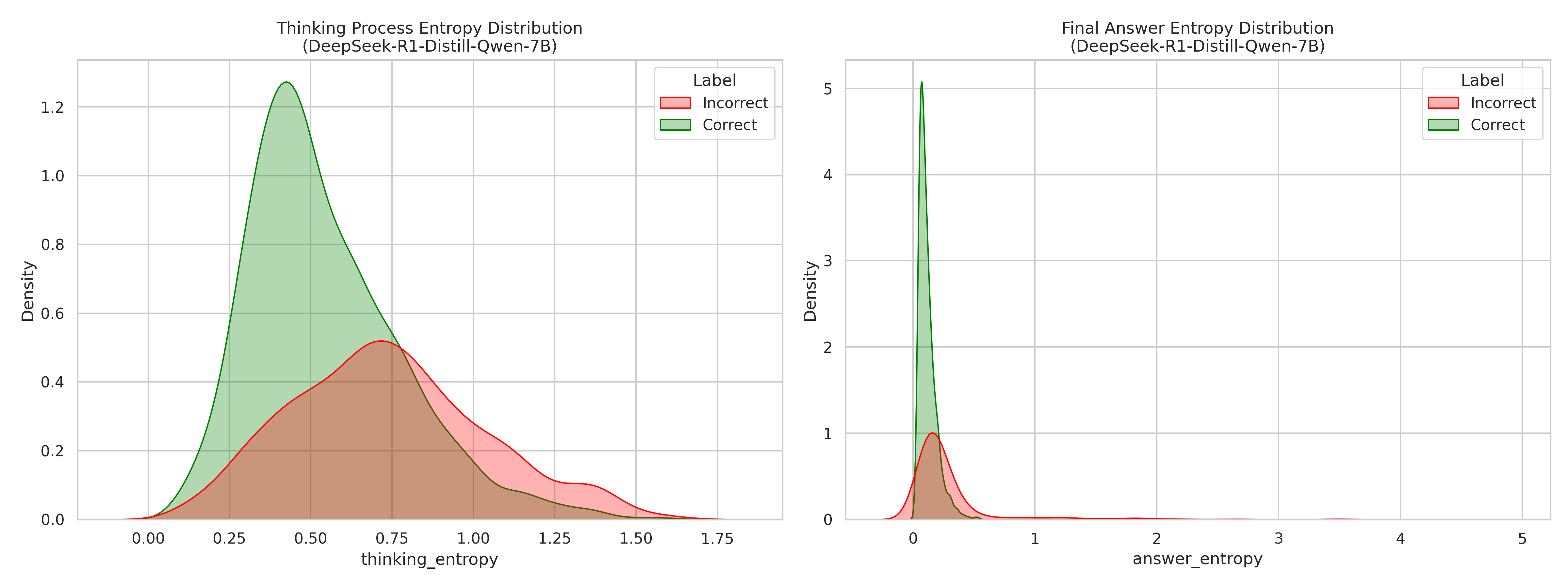}
\vspace{-1mm}
\caption{Entropy distributions for DeepSeek-R1-Distill-Qwen-7B (thinking-side vs.\ answer-side; correct vs.\ incorrect).}
\label{fig:app_ds_7b_dist}
\vspace{-2mm}
\end{figure*}

\noindent\textbf{Entropy distributions.}
Figures~\ref{fig:app_ds_1p5b_dist} and \ref{fig:app_ds_7b_dist} visualize kernel density estimates of entropy for correct and incorrect rollouts.
Answer-side entropy exhibits a clearer separation: correct rollouts concentrate in low-entropy regions, whereas incorrect rollouts shift toward higher entropy, consistent with the AUC comparison in Table~\ref{tab:b1_summary}.

\begin{figure*}[!htbp]
\centering
\begin{subfigure}[t]{0.5\textwidth}
  \centering
  \includegraphics[width=\linewidth]{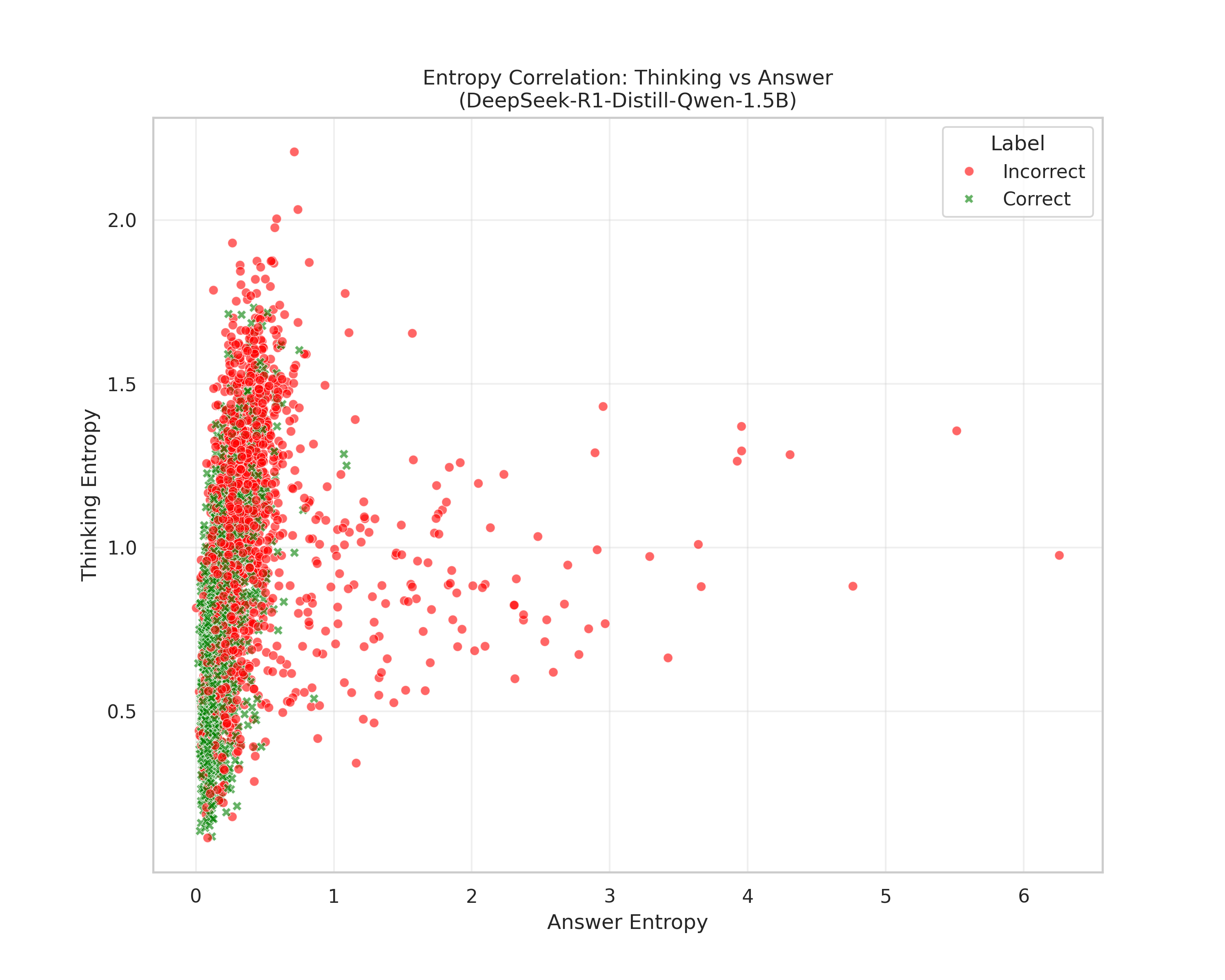}
  \caption{Thinking-side entropy vs.\ answer-side entropy (DeepSeek-R1-Distill-Qwen-1.5B).}
\end{subfigure}\hfill
\begin{subfigure}[t]{0.5\textwidth}
  \centering
  \includegraphics[width=\linewidth]{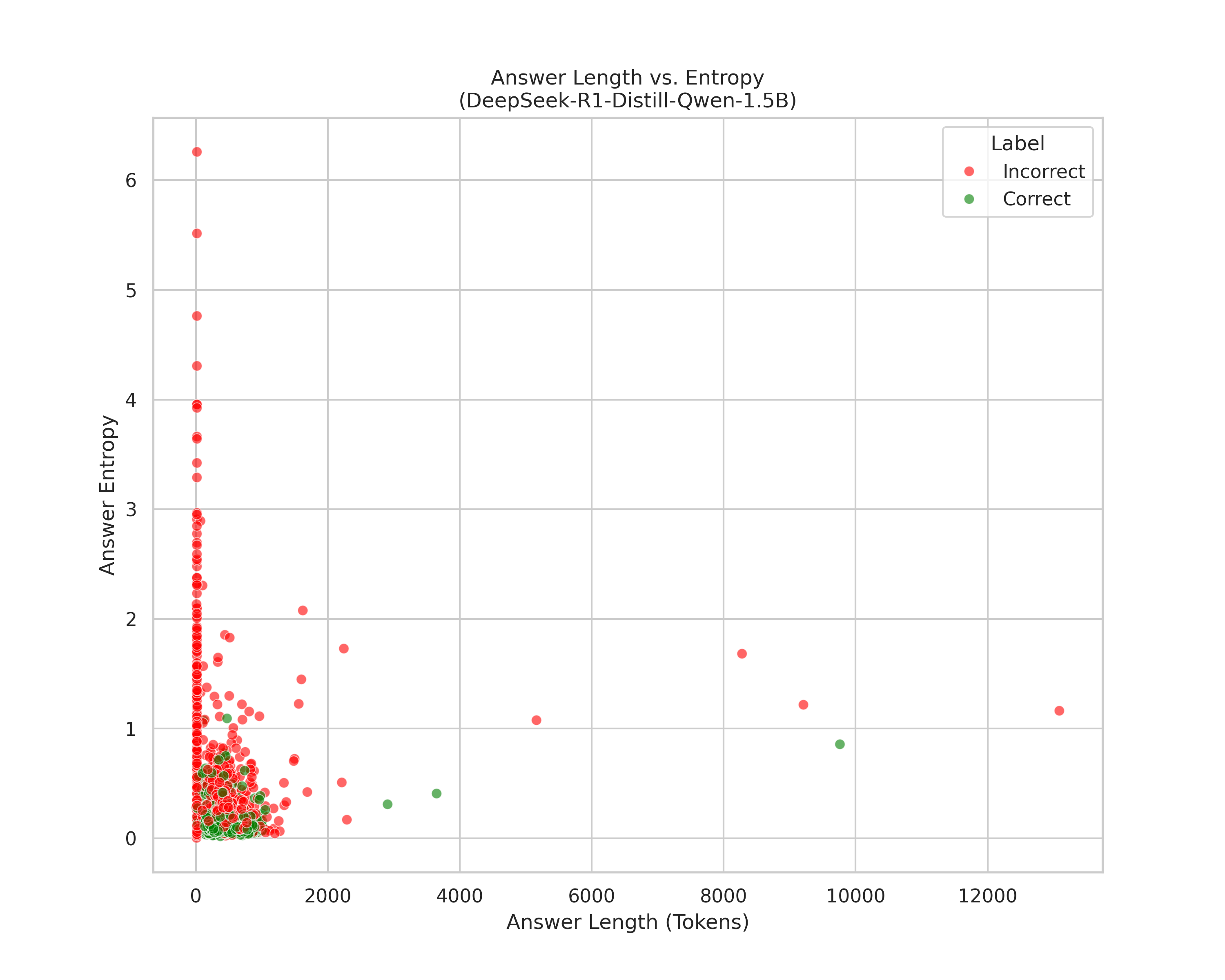}
  \caption{Answer length vs.\ answer-side entropy (DeepSeek-R1-Distill-Qwen-1.5B).}
\end{subfigure}
\vspace{-1mm}
\caption{Supplementary scatter plots for DeepSeek-R1-Distill-Qwen-1.5B.}
\label{fig:app_ds_1p5b_scatter}
\vspace{-2mm}
\end{figure*}

\begin{figure*}[!htbp]
\centering
\begin{subfigure}[t]{0.5\textwidth}
  \centering
  \includegraphics[width=\linewidth]{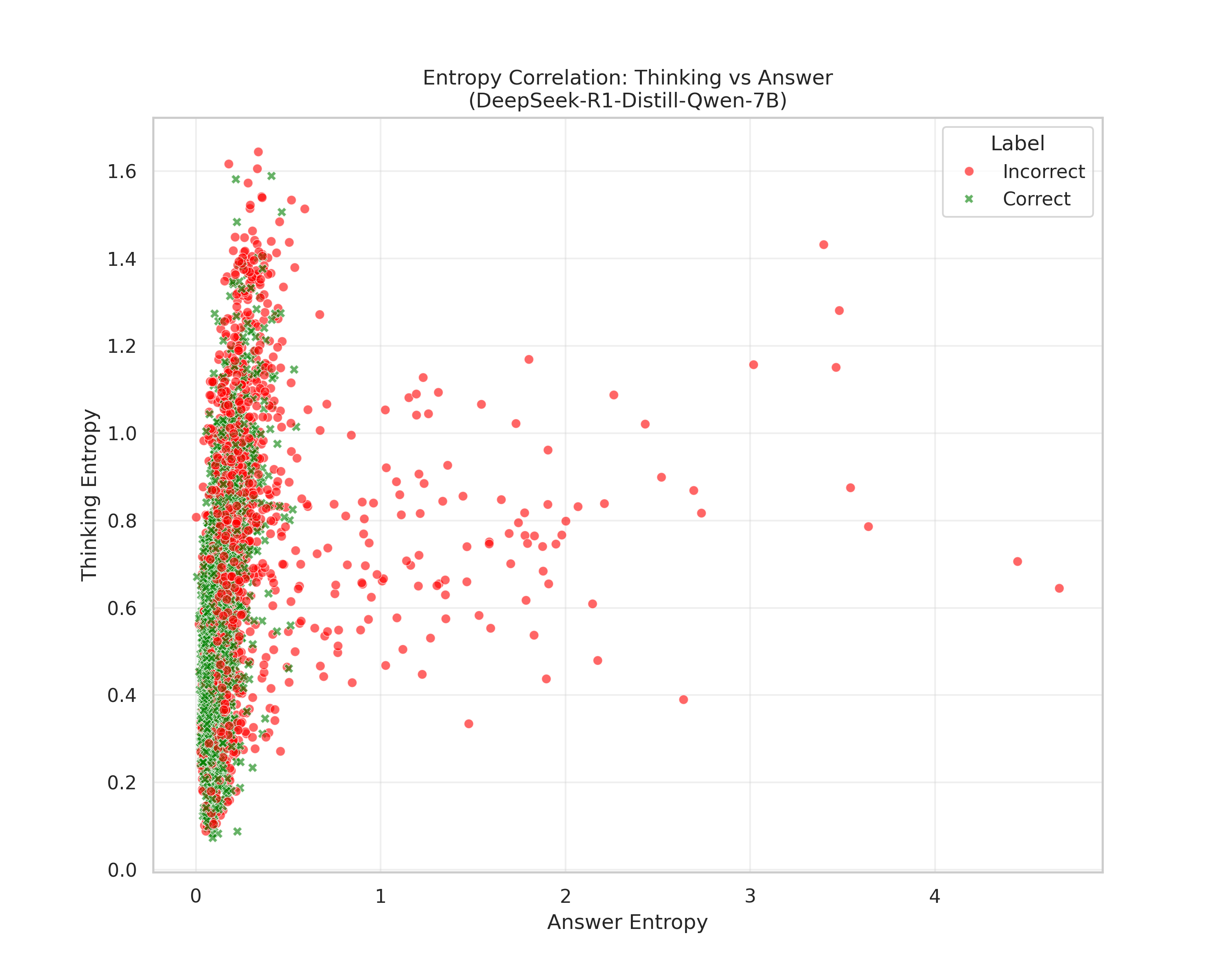}
  \caption{Thinking-side entropy vs.\ answer-side entropy (DeepSeek-R1-Distill-Qwen-7B).}
\end{subfigure}\hfill
\begin{subfigure}[t]{0.5\textwidth}
  \centering
  \includegraphics[width=\linewidth]{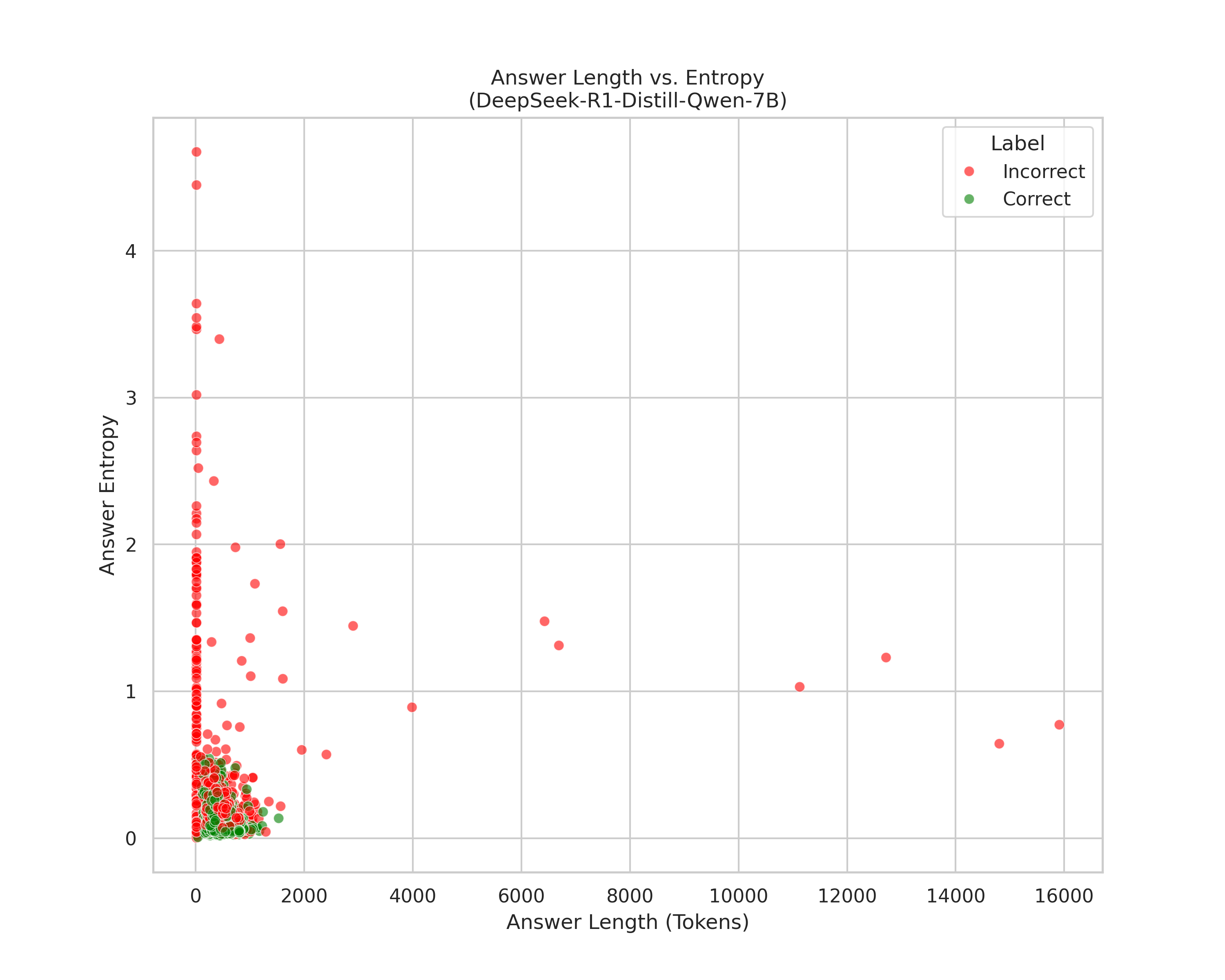}
  \caption{Answer length vs.\ answer-side entropy (DeepSeek-R1-Distill-Qwen-7B).}
\end{subfigure}
\vspace{-1mm}
\caption{Supplementary scatter plots for DeepSeek-R1-Distill-Qwen-7B.}
\label{fig:app_ds_7b_scatter}
\vspace{-2mm}
\end{figure*}

\noindent\textbf{Not a length proxy.}
We plot \textit{answer length} versus \textit{answer-side entropy} in Fig.~\ref{fig:app_ds_1p5b_scatter} and Fig.~\ref{fig:app_ds_7b_scatter}.
We observe no strong monotonic relationship between length and entropy; incorrect rollouts tend to lie above correct ones in entropy at comparable lengths, indicating that the proxy captures content-level uncertainty rather than token-count artifacts.

\begin{figure*}[!htbp]
\centering
\begin{subfigure}[t]{0.48\textwidth}
  \centering
  \includegraphics[width=\linewidth]{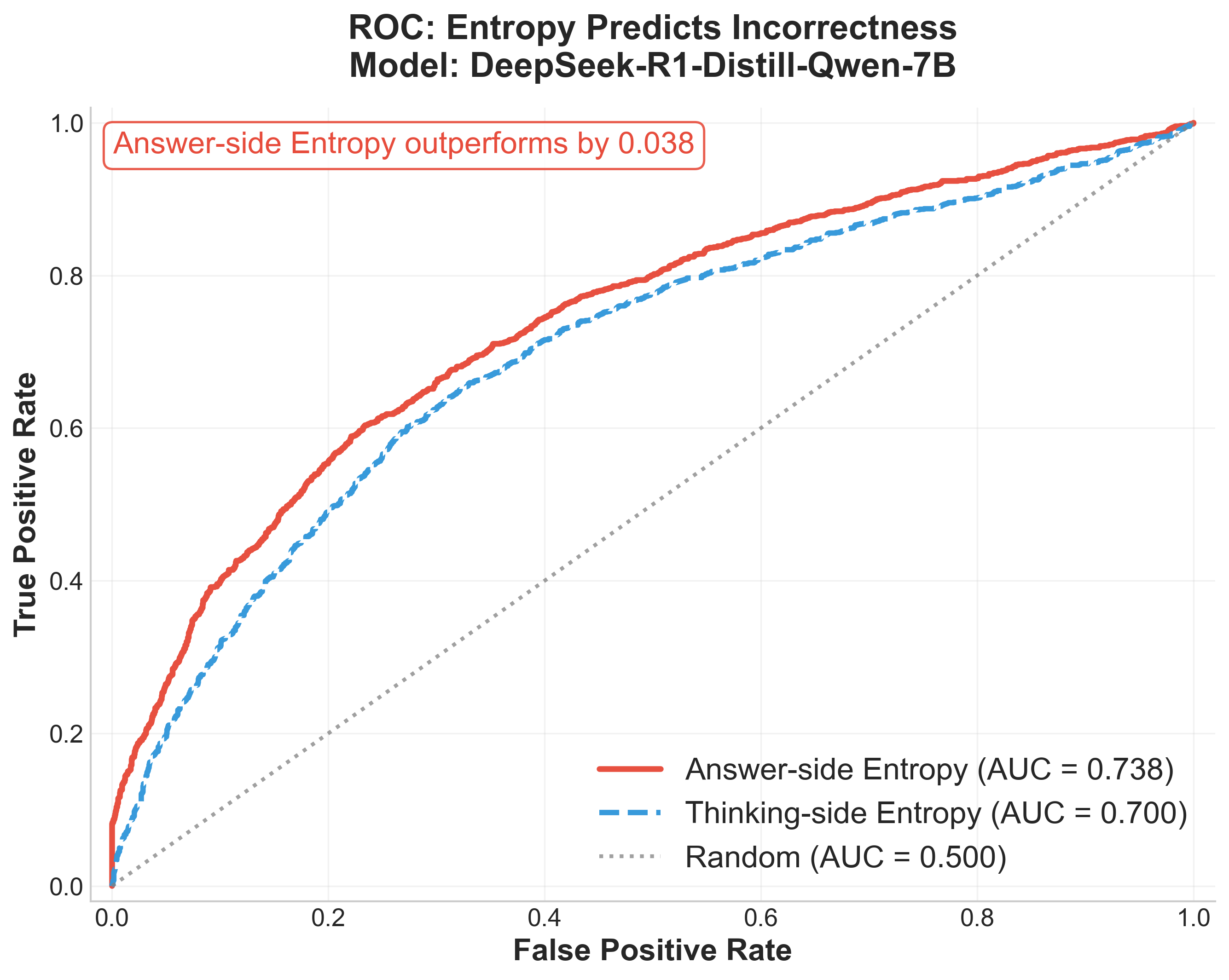}
  \caption{DeepSeek-R1-Distill-Qwen-7B}
  \label{fig:roc_curve}
\end{subfigure}\hfill
\begin{subfigure}[t]{0.48\textwidth}
  \centering
  \includegraphics[width=\linewidth]{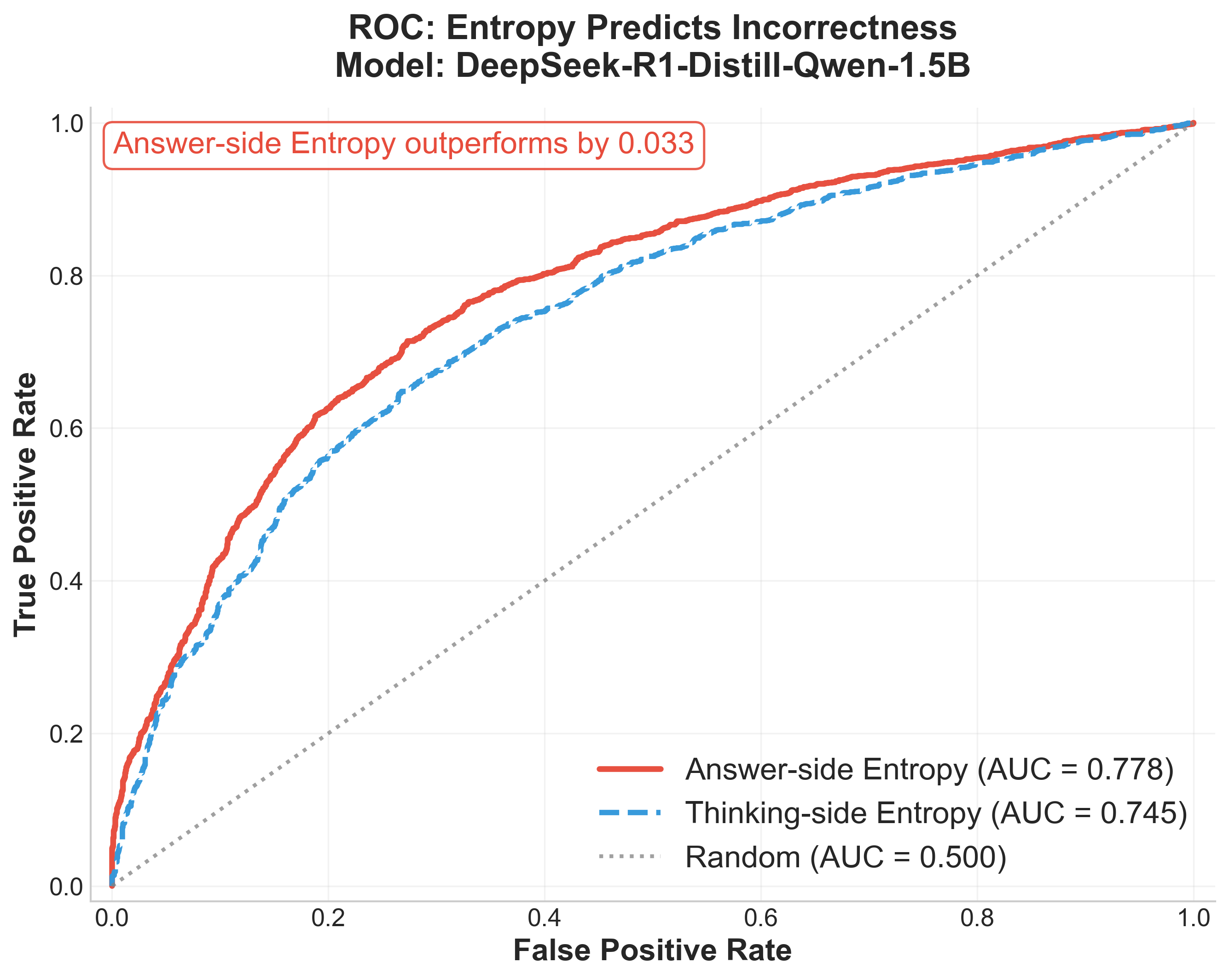}
  \caption{DeepSeek-R1-Distill-Qwen-1.5B}
  \label{fig:app_roc_deepseek_1p5b}
\end{subfigure}
\caption{ROC diagnostics for predicting incorrect rollouts using entropy on DeepSeek-R1-Distill-Qwen backbones, comparing thinking-side vs.\ answer-side entropy.}
\label{fig:app_roc_deepseek}
\end{figure*}

\noindent\textbf{ROC diagnostics.}
Fig.~\ref{fig:app_roc_deepseek_1p5b} and Fig.~\ref{fig:roc_curve} report ROC curves for predicting incorrectness using entropy signals.
Answer-side entropy consistently achieves higher TPR under the same FPR than thinking-side entropy.

\noindent\textbf{Supplementary plots.}
We further visualize:
(i) entropy distributions for correct vs.\ incorrect samples,
(ii) the relationship between thinking-side and answer-side entropy,
and (iii) answer length vs.\ answer-side entropy.

\vspace{-1mm}

\vspace{-1mm}

\vspace{-1mm}

\vspace{-1mm}

\end{document}